\documentclass{article}

 \usepackage[preprint]{neurips_2026}

\usepackage[utf8]{inputenc} 
\usepackage[T1]{fontenc}    
\usepackage{hyperref}       
\usepackage{url}            
\usepackage{booktabs}       
\usepackage{amsfonts}       
\usepackage{nicefrac}       
\usepackage{microtype}      
\usepackage{graphicx} 
\usepackage{amsmath} 
\usepackage{listings}
\usepackage{xcolor}
\usepackage{wrapfig}
\usepackage{multirow}
\usepackage{booktabs}
\usepackage{fontawesome5}
\usepackage{placeins}
\newlength{\panelheight}
\usepackage[most]{tcolorbox}
\tcbuselibrary{breakable,listings,skins}

\usepackage{tcolorbox}
\tcbuselibrary{listings, breakable, skins}

\newtcblisting{promptbox}[1]{
  colback=gray!5!white,       
  colframe=gray!75!black,     
  fonttitle=\bfseries\sffamily, 
  title=#1,                   
  breakable,                  
  enhanced,
  listing only,               
  listing options={
    basicstyle=\ttfamily\scriptsize, 
    breaklines=true,          
    columns=fullflexible,
    showstringspaces=false
  },
  left=2.5mm, right=2.5mm, top=1.5mm, bottom=1.5mm 
}

\usepackage[table]{xcolor}
\usepackage{array}

\definecolor{tacobase}{RGB}{102,204,102}   
\definecolor{tacodark}{RGB}{0,102,0}       
\definecolor{tokdark}{RGB}{85,128,85}      
\definecolor{mutedgray}{RGB}{120,120,120}  

\newcommand{\gaincell}[3]{%
  \cellcolor{tacobase!#3}\textbf{#1}{\scriptsize\textcolor{tacodark}{$^{\uparrow #2}$}}%
}

\newcommand{\tokcell}[3]{%
  \cellcolor{tacobase!#3}\textbf{#1}{\scriptsize\textcolor{tokdark}{$^{\downarrow #2}$}}%
}

\newcommand{\tokup}[2]{%
  #1{\scriptsize\textcolor{mutedgray}{$^{\uparrow #2}$}}%
}
\title{A Self-Evolving Framework for Efficient Terminal Agents via Observational Context Compression}

\newcommand{\equalcontrib}{\textsuperscript{*}}
\newcommand{\corecontrib}{\textsuperscript{\;\textdagger}}
\newcommand{\corresponding}{\textsuperscript{\;\textdaggerdbl}}

\author{%
\makebox[\textwidth][c]{%
    \textbf{Jincheng Ren}\textsuperscript{1,2}\equalcontrib\quad
    \textbf{Siwei Wu}\textsuperscript{1}\equalcontrib\corecontrib\quad
    \textbf{Yizhi Li}\textsuperscript{1}\equalcontrib\quad
    \textbf{Kang Zhu}\textsuperscript{2}\quad
    \textbf{Shu Xu}\textsuperscript{4}\quad
    \textbf{Boyu Feng}\textsuperscript{2}\quad
    \textbf{Ruibin Yuan}\textsuperscript{5}%
}\\
\makebox[\textwidth][c]{%
    \textbf{Wei Zhang}\textsuperscript{6}\quad
    \textbf{Riza Batista-Navarro}\textsuperscript{1}\quad
    \textbf{Jian Yang}\textsuperscript{6}\corresponding\quad
    \textbf{Chenghua Lin}\textsuperscript{1}\corresponding%
}\\
\makebox[\textwidth][c]{%
    \textsuperscript{1}University of Manchester\quad
    \textsuperscript{2}MAP\quad
    \textsuperscript{4}HKUST(GZ)\quad
    \textsuperscript{5}HKUST\quad
    \textsuperscript{6}Beihang University%
}
}
\begin{document}

\maketitle

\begingroup

\renewcommand{\thefootnote}{\fnsymbol{footnote}}

\footnotetext[1]{Equal contribution.}
\footnotetext[2]{Core idea, code guidance, and main writing.}
\footnotetext[3]{Corresponding authors.}

\endgroup

\newcommand{\agentname}{TACO}

\begin{abstract}
Terminal observations are not ordinary long-context text: they are heterogeneous, low-information-density execution traces in which sparse but exact evidence (e.g., error messages and file paths) is interleaved with large amounts of redundant terminal output. For long-horizon CLI agents, retaining raw observations rapidly increases context cost and can dilute critical signals, while LLM-based summarization or fixed heuristics often fail to adapt across heterogeneous terminal tasks and may discard precise task-relevant evidence.
We propose TACO, the first self-evolving \textbf{T}erminal \textbf{A}gent \textbf{C}ompressi\textbf{o}n framework, which treats compression rules as reusable, preservation-aware knowledge acquired from interaction trajectories.
Rather than relying on manually designed rules, static pruning, or task-specific compressor training, \agentname{} autonomously discovers, refines, and reuses structured compression rules from agent interaction trajectories. Our global rule pool mechanism accumulates effective rules across tasks, while intra-task rule evolution adapts them online to the current workflow. This allows \agentname{} to filter redundant terminal observations while preserving exact task-relevant evidence.
Across \textbf{six} benchmarks, including \textbf{TB~1.0},
\textbf{TB~2.0}, \textbf{SWE-Bench Lite}, \textbf{CompileBench}, \textbf{DevEval}, and \textbf{CRUST-Bench}, \agentname{} consistently maintains or improves task success across models and agent scaffolds. On TerminalBench, \agentname{} yields \textbf{1\%--4\%} absolute accuracy gains under standard evaluation and improves accuracy by \textbf{2\%--3\%} under matched token budgets. On downstream benchmarks, it reduces total token consumption by \textbf{12\%--27\%} while maintaining or improving task success. These results show that self-evolving observation compression can unlock latent capability in existing CLI agents by allocating context budget toward task-relevant evidence, without model fine-tuning or human-crafted compression rules. The code for \agentname{} is available at \url{https://github.com/multimodal-art-projection/TACO.git}.
\end{abstract}

\section{Introduction}

Recent advances in code foundation models, code intelligence, and agentic code systems have enabled increasingly capable software-engineering agents~\citep{yang2025code,yang2026incoder,yang2026incoder_thinking,yang2026iquest}, yet terminal-centric tasks such as repository debugging, compilation, testing, and environment interaction remain challenging. Unlike short-form code generation, CLI agents operate through long-horizon interaction loops: they execute commands, observe terminal outputs, and condition future actions on accumulated environmental feedback~\citep{merrill2026terminal,jimenez2024swebench}. As these histories grow, mainstream terminal agents often preserve accumulated raw command outputs to avoid losing useful feedback, creating a central bottleneck for both efficiency and reliability.

\begin{wrapfigure}{r}{0.5\columnwidth}
    \centering
    \vspace{-0.8em}
    \includegraphics[width=0.50\columnwidth]{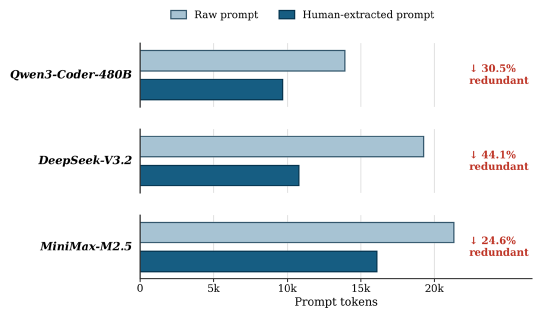}
    \caption{Effective-text extraction reveals substantial removable redundancy in terminal-agent histories on TB 2.0.}
    \label{fig:motivation}
    \vspace{-1.0em}
\end{wrapfigure}

Our trajectory analysis shows that this bottleneck is not merely caused by long context, but by the highly uneven information density of terminal observations. As shown in Fig.~\ref{fig:motivation}, manually extracting effective text from 50 sampled TB~2.0 trajectories removes 24.6\%--44.1\% of raw prompt tokens across the evaluated models, indicating that a substantial portion of terminal-agent histories is occupied by low-value content rather than actionable information. However, this redundancy is not cleanly separable from useful content: verbose logs, installation traces, build outputs, and test reports may contain exact evidence required for future decisions, including error messages, file paths, failing test names, command arguments, and build targets. This suggests that terminal-observation compression should not simply maximize shortening; it must perform preservation-aware filtering, selectively removing low-value terminal output while preserving the exact evidence needed for future actions.

\textbf{Existing compression methods are not designed for preservation-aware terminal observation filtering.}
Generic LLM summarization is flexible, but it may paraphrase, omit, or blur exact terminal signals, which can be harmful when later actions depend on precise strings such as error messages, file paths, or test names.
Static heuristics and expert-crafted rules can preserve such signals when carefully designed, but they are brittle across commands, repositories, languages, and task domains, and require substantial manual engineering.
Training-based methods such as SWE-Pruner~\citep{wang2026swe} provide more adaptive pruning, but they require additional training data and are mainly optimized for SWE-Bench-style software-engineering workflows, limiting their applicability to broader terminal environments.
These limitations raise a natural question: \emph{can a terminal agent autonomously discover reusable compression knowledge from its own interaction trajectories, without human-crafted rules or compressor training?}

To address this problem, we propose \agentname{}, a plug-and-play, unsupervised test-time adaptation \textbf{T}erminal \textbf{A}gent \textbf{C}ompressi\textbf{o}n framework.
\agentname{} reframes terminal-observation compression from a one-shot shortening operation into a continual knowledge-acquisition process over terminal-output patterns. Instead of treating compression rules as fixed heuristics or task-specific parameters, \agentname{} treats them as reusable preservation-aware knowledge that can be discovered, refined, and transferred across workflows.
During execution, an LLM proposes structured rules for observation patterns, and a conservative executor applies them only when their triggers match while preserving critical evidence by design. Within each task, \agentname{} adapts the active rule set using implicit feedback from the agent, such as requests for full output or repeated commands indicating over-compression. Across tasks, effective rules are accumulated, scored, and reused through a Global Rule Pool, allowing compression knowledge discovered in one workflow to benefit subsequent tasks.

We integrate \agentname{} into mainstream agent frameworks and evaluate it across multiple terminal-related benchmarks with strong backbone models. In summary, our main contributions are as follows:






\begin{enumerate}

    \item \textbf{We introduce self-evolving terminal-observation compression for CLI agents.}
    We formulate terminal-agent context compression as a self-evolving rule-learning problem, where terminal-output patterns are converted into reusable, preservation-aware compression rules. \agentname{} realizes this idea as a \textbf{plug-and-play}, \textbf{unsupervised test-time adaptation}, and \textbf{human-free} framework that autonomously discovers, refines, and reuses structured rules from agent interaction trajectories, without task-specific compressor training or human-crafted rules.

    \item \textbf{\agentname{} improves the accuracy--token trade-off on TerminalBench.}
    On TerminalBench 1.0 and 2.0, integrating \agentname{} into existing agent frameworks yields consistent \textbf{1\%--4\%} absolute accuracy gains across strong backbone models. 
    Under matched token budgets, \agentname{} further improves accuracy by around \textbf{2\%--3\%}, showing that it helps agents use the same context budget more effectively.

    \item \textbf{\agentname{} adapts across heterogeneous terminal benchmarks.}
    Beyond TerminalBench, we evaluate \agentname{} on SWE-Bench Lite, CompileBench, DevEval, and CRUST-Bench, where it maintains or improves task success while reducing total token consumption by \textbf{12\%--27\%}.
    These results show that \agentname{} can be directly integrated into agents for diverse terminal-related benchmarks and automatically adapt its compression rules to different terminal tasks, without task-specific training or manually designed rules.

\end{enumerate}

\section{Self-evolving Compression Agentic Framework}
\label{sec:method}
 
\begin{figure*}[t]
    \centering
    \includegraphics[width=0.90\textwidth]{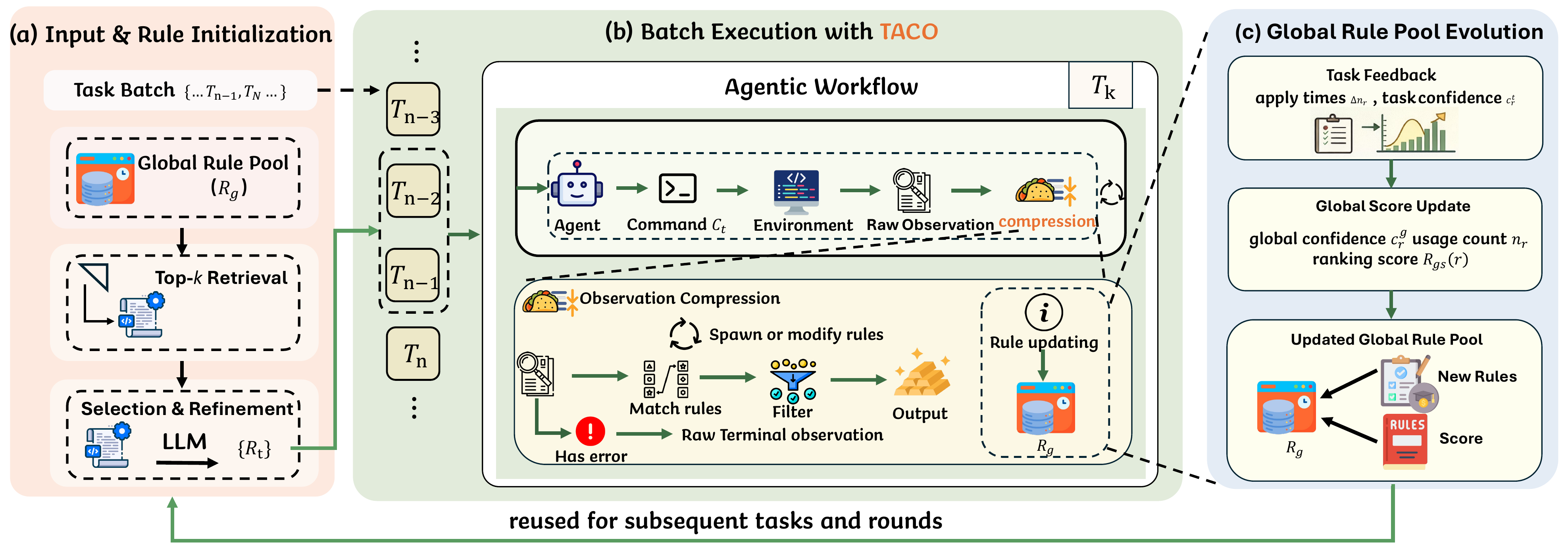}
    \caption{
    Overview of \agentname{}.
    }
    \label{fig:taco_main}
    \vspace{-1.5em}
    \end{figure*}

During a terminal agent's execution loop, each step receives an environment observation that often contains substantial redundancy. To improve token efficiency, we propose TACO, a self-evolving context compression framework that compresses terminal observations with a dynamically evolving set of compression rules. Each rule specifies when and how an observation should be compressed.
As shown in Fig.~\ref{fig:taco_main}, TACO consists of three components: (1) \textbf{Terminal Observation Compression} (top right), which applies evolved rules at each step; (2) \textbf{Intra-Task Rule Set Evolution} (middle right), which updates rules online from compression outcomes within the current task; and (3) \textbf{Global Rule Pool Evolution} (bottom right), which refines and shares effective rules across tasks.



\subsection{Task Definition}
Given a terminal task $T$, an agent $A$ iteratively interacts with the environment. At each step $t$, the agent generates a command $C_{t+1} = A(T, S_t)$ conditioned on the history sequence $S_t$ of preceding command-observation pairs, and receives a raw terminal observation $O_{t+1}$. This repeats until task completion. For efficiency, we process \(N\) tasks in parallel, where \(N\) is the batch size.

\subsection{TACO Adapter}
\textbf{TACO} is a plug-and-play terminal observation compression adapter that can be integrated into different terminal-agent scaffolds. Specifically, after the host agent executes a command $C_t$ and receives the raw terminal observation $O_t$, TACO returns a compressed observation $\tilde{O}_t$ according to the task-specific rule set $R_t$. 
The detailed compression process is described in Sec.~\ref{sec:terminal_output_compression}.


\subsection{Rule Initialization}
\label{sec: Rule Initialization}

\paragraph{Rule Definition.} A rule consists of an applicability condition together with compression parameters, such as trigger patterns, retained patterns, removed patterns, and conservative retention bounds. During execution, these structured rules are instantiated by a fixed rule executor into concrete filtering behavior. This design constrains self-evolution to a safe and reusable rule space, enabling TACO to remove low-value noise while preserving critical information.  Detailed rule schema and execution examples are provided in Appendix~\ref{Rule Format and Examples}.

\paragraph{Global Rule Pool.}
Since many compression patterns discovered during task execution are reusable across tasks, such as compressing \texttt{pip install} progress in package-installation scenarios, we maintain a Global Rule Pool \(R_g\) that accumulates structured compression rules, with each rule \(r \in R_g\) initialized with a global confidence score \(c_r^{g}=1.0\) for tracking reliability and a ranking score \(R_{gs}(r)\) for global ranking and task-level retrieval. After a task is completed, effective rules are selectively written back to the pool, with their global scores initialized or updated based strictly on their empirical performance during the task.




\paragraph{Task-Level Rule Selection.}
Different tasks may require distinct compression behaviors.
We first retrieve the top-$k$ rules from \(R_g\) according to their ranking scores \(R_{gs}(r)\) as candidates.
Given the task description and objective, an LLM then selects and refines these candidates to better align with the current context.
This task-conditioned selection prevents high-ranked but irrelevant rules from being activated: candidates whose triggers do not match the current task are filtered out regardless of their \(R_{gs}(r)\) values.
The resulting rules initialize the task-specific active rule set \(R_t\).
Additional implementation details are provided in Appendix~\ref{prompt}.

\subsection{Terminal Observation Compression}
\label{sec:terminal_output_compression}

At each interaction step, the agent executes a terminal command and receives a raw terminal observation $O_t$. 
Observations containing explicit error or failure signals, such as syntax errors and exception traces, are treated as \textbf{Critical}, as they often inform subsequent decisions.
TACO leaves such outputs unchanged and applies the active task-specific rules in $R_t$ only to non-Critical outputs:
\begin{equation}
\tilde{O}_t =
\begin{cases}
O_t, & \text{if } O_t \text{ is \textbf{Critical}},\\
F_{R_t}(O_t \mid C_t), & \text{otherwise},
\end{cases}
\end{equation}
where $F_{R_t}$ denotes a conservative rule-based compression operator induced by the active rule set $R_t$.

\subsection{Intra-Task Rule Set Evolution}

When solving each task, TACO updates the task-specific rule set dynamically.

\paragraph{Adding Rules}
When a terminal observation is not handled by any rule in $R_t$, TACO treats it as an uncovered output and invokes an LLM to generate a new rule $r$, which is then added to $R_t$ for subsequent steps.


\paragraph{Updating Rules}
TACO updates rules based on implicit feedback from the agent’s behavior after consuming the compressed observation $\tilde{O}_t$. Behaviors such as requesting the full output, indicating missing critical details, or repeating the same command to recover more information are treated as over-compression complaints. In response, TACO traces back the rules that triggered and modified the current observation, suppresses their use for subsequent steps, and injects more conservative replacement variants via LLMs. The corresponding prompt is provided in Appendix~\ref{prompt}.

\subsection{Global Rule Pool Evolution}
\label{subsec:Global Rule Pool Evolution}
 
To accumulate reusable compression knowledge across tasks, TACO writes effective task-level rules back to the Global Rule Pool and updates their global statistics according to their observed behavior in the completed task.

\paragraph{Global Rule Pool Updating.}
After completing a task, \agentname{} updates the Global Rule Pool based on task-level evidence collected from \(R_t\). 
For each rule \(r \in R_t\), we record two quantities: \(\Delta n_r\), the number of successful applications of \(r\) in the current task, and \(c_r^{t} \in [0,1]\), its final task-level confidence. 
A rule is written back to \(R_g\) only if it has been successfully applied at least once and remains reliable, i.e., \(\Delta n_r \geq 1\) and \(c_r^{t} \geq \tau\), where \(\tau\) is a small confidence threshold. 
Rules that receive explicit complaints are assigned \(c_r^{t}=0\) for the current task and are therefore excluded from write-back. 
If such a rule already exists in \(R_g\), its global confidence \(c_r^{g}\) is further decayed, lowering its ranking score and reducing its chance of being retrieved in subsequent tasks.



\paragraph{Global Rule Score Updating.}
For a newly discovered effective rule, TACO initializes its global confidence and usage statistics directly from the current task. For an existing effective rule, TACO updates its global confidence $c_r^{g}$ using the task-end confidence $c_r^{t}$ and updates its cumulative usage count with the newly observed successful applications.

Each rule $r$ in the Global Rule Pool is assigned a ranking score:
\begin{equation}
R_{gs}(r) = c_r^{g} \cdot (n_r + 1),
\label{eq:global_ranking_score}
\end{equation}
where $c_r^{g}$ denotes its global confidence and $n_r$ denotes its cumulative number of successful applications across tasks. 
This score prioritizes rules that are both reliable and broadly reusable: complaints reduce $c_r^{g}$ and thus down-rank problematic rules, while frequent successful use increases $n_r$. 
Newly generated rules are initialized with $c_r^{g}=1.0$, so they are not prematurely excluded before accumulating usage evidence. 
We use this score only to retrieve the top-$k$ candidates, after which the LLM performs task-conditioned selection to decide which rules to activate.

\paragraph{Multi-round Evolution and Batch Size.} 

To obtain broader and more transferable compression rules, TACO performs self-evolution across tasks rather than relying only on evolution within a single task. Moreover, we run \textbf{multiple evolution rounds} on the same dataset until convergence is reached. Under parallel execution, effective rules from completed tasks are written back to the Global Rule Pool and reused to initialize later tasks. As a result, the batch size $N$ affects the speed of rule propagation and can influence final performance, which we analyze in the Appendix~\ref{subsec:Hyperparameter Selection}.

\subsection{Reward-Free Unsupervised Rule Evolution Protocol}
\label{sec:frontier}


\paragraph{Algorithm Convergence.} Because our self-evolving method continuously acquires new rules, the update dynamics of the Global Rule Pool provide a natural convergence signal. Therefore, we use the change rate of the Top-$K$ rules in the Global Rule Pool to assess convergence. Specifically, we define a \textbf{reward-free} metric, \emph{Retention}, which measures the proportion of rules that remain in the Top-$K$ after one round of evolution on the dataset. It is calculated as follows:
\vspace{-0.2em}
\begin{equation}
\mathrm{Retention}^{(i)}_K
=
\frac{
\left|
\mathrm{TopK}\!\left(R_g^{(i-1)}\right)
\cap
\mathrm{TopK}\!\left(R_g^{(i)}\right)
\right|
}{K}\times 100\%,
\end{equation}
\vspace{-0.2em}
where $R_g^{(i)}$ is the Global Rule Pool after the $i$-th run.
A higher Retention value indicates that the effective rule frontier has become more stable. In this work, we set $K=30$. Additional discussion on the choice of $K$ is provided in the Appendix~\ref{subsec:Hyperparameter Selection}.


\paragraph{Reward-Free Rule Evolution and Leakage Prevention.}
As detailed in Appendix~\ref{appdix:Leakage}'s Tab.~\ref{tab:leakage_protocol}, TACO evolves compression rules solely from terminal observations, without accessing benchmark answers, hidden tests, verifier outcomes, task success labels, or leaderboard feedback.
Rule updates depend only on observation-level signals, including rule triggers, application counts, terminal observation, and subsequent interactions that indicate possible over-compression.
The stopping criterion is also \textbf{reward-free}: evolution terminates when the metric \emph{Retention} stabilizes.

\section{Experiment Setup}
\label{Sec: Experiment Setup}

\subsection{Benchmarks}

To validate the effectiveness and generality of \agentname{}, we not only evaluate its performance gains with various agentic backbones on TerminalBench (TB~1.0 and TB~2.0), but also conduct experiments on a range of terminal-related benchmarks (i.e., SWE-Bench Lite~\citep{jimenez2024swebench}, CompileBench, DevEval~\citep{li2024deveval}, CRUST-Bench~\citep{khatry2025crustbenchcomprehensivebenchmarkctosaferust}).

\subsection{LLMs and Agent Scaffolds}

To evaluate the generality of \agentname{} across both model families and agent implementations, we instantiate our method with multiple backbone LLMs and evaluate it on different agent scaffolds.

\paragraph{LLMs.}
We consider closed-source and open-source models.
For closed-source models, we evaluate the Claude, GPT, and Gemini series, among the strongest in coding capability.
For open-source models, beyond state-of-the-art models on terminal tasks such as MiniMax, DeepSeek~\citep{liu2025deepseek}, GLM~\citep{5team2025glm45agenticreasoningcoding}, Kimi-K2-Instruct~\citep{kimiteam2025kimik2openagentic}, and Nex-N1~\citep{cai2025nex} series, we evaluate several Qwen3-series models with fewer than 40B parameters.


\paragraph{Agent Scaffolds.}
We use two representative agent scaffolds. 
\textbf{Terminus-2}, introduced by TerminalBench~\citep{merrill2026terminal}, has been adapted to several terminal-oriented benchmarks, and we use it as the baseline scaffold for TB, CompileBench, DevEval, and CRUST-Bench. 
\textbf{Mini-SWE-Agent} is a minimalist software-engineering agent for repository-level issue solving and command-line interaction; following SWE-Bench~\citep{yang2024swe}, we use it for SWE-Bench Lite.

\paragraph{Evaluation protocol and token accounting.}
Unless otherwise specified, all in-house results are averaged over five
evaluation runs. All reported token counts are end-to-end totals,
including both backbone-agent execution tokens and auxiliary LLM calls used by
\agentname{}. We report run-level standard deviations for the main accuracy
results in Appendix~\ref{metrics_details}.

\section{Results and Discussion}
\label{sec:Results and Discussion}

\subsection{Adaptation Across Heterogeneous Terminal Benchmarks}
\label{Generalization Across Terminal and Benchmarks}

\begin{wraptable}{r}{0.55\textwidth}
\vspace{-1.0em}
\centering
\caption{Results across six benchmarks.}
\label{tab:generability_exp}
\scriptsize     
\setlength{\tabcolsep}{3.6pt}
\renewcommand{\arraystretch}{1.05}
\begin{tabular}{@{}lcccc@{}}
\toprule
\multirow{2}{*}{\textbf{Benchmark}} 
& \multicolumn{2}{c}{\textbf{Accuracy (\%)}} 
& \multicolumn{2}{c}{\textbf{Total Tokens (millions)}} \\
\cmidrule(lr){2-3}\cmidrule(lr){4-5}
& Baseline & \agentname{} & Baseline & \agentname{} \\
\midrule
SWE-Bench-Lite & 56.30 & \gaincell{57.12}{0.82}{12} & 307.61 & \tokcell{270.53}{12.1\%}{32} \\
CompileBench   & 75.00 & 75.00                      & 14.55  & \tokcell{11.41}{21.6\%}{45} \\
DevEval        & 38.10 & \gaincell{39.74}{1.64}{24}    & 36.72  & \tokcell{26.82}{27.0\%}{55} \\
CRUST-Bench    & 47.00 & \gaincell{48.05}{1.05}{16}    & 163.53 & \tokcell{134.97}{17.5\%}{40} \\
TB~1.0         & 42.30 & \gaincell{45.25}{2.95}{36}    & 29.74  & \tokcell{23.43}{21.2\%}{45} \\
TB~2.0         & 42.80 & \gaincell{44.16}{1.36}{20}    & 113.74 & \tokcell{110.63}{2.7\%}{14} \\
\bottomrule
\end{tabular}
\vspace{-1.6em}
\end{wraptable}


To evaluate the adaptation ability of \agentname{} across heterogeneous terminal
environments, we conduct experiments on several terminal-related benchmarks, including TerminalBench (TB~1.0 and TB~2.0), SWE-Bench Lite, CompileBench, DevEval, and CRUST-Bench. We use MiniMax-M2.5 as the backbone model for these evaluations. As shown in Tab.~\ref{tab:generability_exp}, \agentname{} consistently maintains or improves task success across all six benchmarks while reducing end-to-end token consumption, especially in log-heavy settings. These results suggest that the self-evolving mechanism of \agentname{} generalizes beyond a single benchmark and adapts to diverse terminal-related environments.
All reported token reductions are measured end-to-end: auxiliary rule-evolution calls are included in total token consumption, yet account for less than 2\% across all benchmarks and about 1\% on average; see Appendix~\ref{app:token_overhead}.

\begin{table*}[t]
\centering
\caption{
Results on TerminalBench (TB) 1.0 and 2.0. 
Shaded cells in the \agentname{} block report gains over the corresponding Terminus-2 baseline using the same backbone.
}
\label{tab:terminal_bench}
\footnotesize
\setlength{\tabcolsep}{5.5pt}
\renewcommand{\arraystretch}{1.05}

\makebox[\textwidth][c]{%
\scalebox{0.91}{%
\begin{tabular}{@{}lclcc@{}}
\toprule
\textbf{Model} & \textbf{Model Size} & \textbf{Agent Scaffold} & \textbf{TB~1.0} & \textbf{TB~2.0} \\
\midrule
\multicolumn{5}{@{}l}{\textit{Open-Source Models ($>$200B)}} \\
\midrule
GLM-4.7                         & 358B & Terminus-2 & 48.75 & 41.00$^{*}$ \\
MiniMax-M2.5                    & 230B & Terminus-2 & 42.30 & 42.80 \\
MiniMax-M2.1                    & 229B & Terminus-2 & 42.50 & 29.20$^{*}$ \\
DeepSeek-V3.2                   & 685B & Terminus-2 & 43.93 & 40.62 \\
DeepSeek-V3.1-Nex-N1            & 685B & OpenHands  & 31.56 & 31.80$^{\dagger}$ \\
Kimi-K2-Instruct                & 1T   & Terminus-2 & 44.59 & 27.80$^{*}$ \\
Qwen3-Coder-480B                & 480B & Terminus-2 & 37.50 & 23.90$^{*}$ \\
Qwen3-235B-A22B-Instruct        & 235B & Terminus-2 & 15.00 & 13.50 \\

\midrule
\multicolumn{5}{@{}l}{\textit{Open-Source Models ($\sim$30B and below)}} \\
\midrule
Qwen3-Coder-30B-A3B-Instruct    & 30B & Terminus-2 & 23.80 & 14.60 \\
Qwen3-30B-A3B-Nex-N1            & 30B & OpenHands  & 25.00 & 8.30$^{\dagger}$ \\
Qwen3-32B-Instruct              & 32B & OpenHands  & 11.25 & 3.40 \\
Qwen3-32B-Instruct              & 32B & Terminus-2 & 11.25 & 3.92 \\
Qwen3-14B-Instruct              & 14B & Terminus-2 & 5.23  & 4.04 \\
Qwen3-8B-Instruct               & 8B  & Terminus-2 & 8.86  & 1.43 \\
Qwen3-32B-Nex-N1                & 32B & OpenHands  & 28.75 & 16.70$^{\dagger}$ \\

\midrule
\multicolumn{5}{@{}l}{\textit{Plugin: \agentname{}}} \\
\midrule
Qwen3-Coder-480B-Instruct
& 480B & \agentname{}+Terminus-2
& \gaincell{38.50}{1.00}{16}
& \gaincell{25.86}{1.96}{25} \\

MiniMax-M2.5
& 230B & \agentname{}+Terminus-2
& \gaincell{45.25}{2.95}{36}
& \gaincell{44.16}{1.36}{20} \\

DeepSeek-V3.2
& 685B & \agentname{}+Terminus-2
& \gaincell{46.25}{2.32}{30}
& \gaincell{42.77}{2.15}{28} \\

Qwen3-32B-Instruct
& 32B & \agentname{}+Terminus-2
& \gaincell{14.13}{2.88}{35}
& \gaincell{7.48}{3.56}{42} \\

Qwen3-14B-Instruct
& 14B & \agentname{}+Terminus-2
& \gaincell{11.25}{6.02}{55}
& \gaincell{6.15}{2.11}{28} \\

Qwen3-8B-Instruct
& 8B & \agentname{}+Terminus-2
& \gaincell{9.22}{0.36}{10}
& \gaincell{3.67}{2.24}{30} \\

\bottomrule
\end{tabular}%
}%
}

\vspace{-1.6em}
\end{table*}

\subsection{Adaptability of \agentname{} Across Agentic Models}

To verify that TACO is not tailored to a specific model, we evaluate it on TerminalBench (TB~1.0 and TB~2.0) across multiple agentic models.

\paragraph{\agentname{} consistently improves agent performance.}
As shown in Tab.~\ref{tab:terminal_bench}, incorporating TACO leads to consistent improvements across the evaluated models on both benchmarks. The absolute gains range from 0.36 to 6.02 points, suggesting that step-wise terminal observation compression can enhance agent performance. In particular, \textbf{Qwen3-Coder-480B} and \textbf{Qwen3-32B-Instruct} improve by \textbf{1.00} and \textbf{2.88} points on \textbf{TB~1.0}, and by \textbf{1.96} and \textbf{3.56} points on \textbf{TB~2.0}, respectively. As discussed in Appendix~\ref{case_study}, we hypothesize that these gains arise from reduced contextual redundancy, which helps the model better focus on task-relevant information. Closed-source TerminalBench results are provided in Appendix~\ref{app:closed_source_tb} for reference.

\begin{wraptable}{r}{0.45\columnwidth}
\centering
\vspace{-1.8em}
\caption{Result with token consumption.
}
\label{tab:compression_efficiency}
\vspace{-0.2em}
\scriptsize
\setlength{\tabcolsep}{2.8pt}
\renewcommand{\arraystretch}{0.96}

\begin{tabular}{@{}lccc@{}}
\toprule
\textbf{Model} & \textbf{Acc. (\%)} & \textbf{Avg Step} & \textbf{Token/Step} \\
\midrule
\multicolumn{4}{@{}l}{\textit{Baseline}} \\
\midrule
Qwen3-Coder-480B & 23.3 & 45.7 & 21,718 \\
DeepSeek-V3.2    & 40.6 & 29.5 & 35,038 \\
MiniMax-M2.5     & 42.8 & 43.2 & 28,631 \\
Qwen3-32B        & 3.9  & 15.7 & 8,472  \\
Qwen3-14B        & 4.0  & 30.3 & 9,663  \\
Qwen3-8B         & 1.4  & 44.3 & 9,579  \\

\midrule
\multicolumn{4}{@{}l}{\textit{+\agentname{}}} \\
\midrule
Qwen3-Coder-480B
& \gaincell{25.8}{2.5}{32}
& 47.0
& \tokcell{19,965}{8.1\%}{38} \\

DeepSeek-V3.2
& \gaincell{42.7}{2.1}{28}
& 30.6
& \tokcell{30,939}{11.7\%}{50} \\

MiniMax-M2.5
& \gaincell{44.1}{1.3}{18}
& 42.6
& \tokcell{28,559}{0.3\%}{8} \\

Qwen3-32B
& \gaincell{7.4}{3.5}{42}
& 19.6
& \tokup{8,735}{3.1\%} \\

Qwen3-14B
& \gaincell{6.1}{2.1}{28}
& 32.4
& \tokcell{9,393}{2.8\%}{18} \\

Qwen3-8B
& \gaincell{3.6}{2.2}{30}
& 68.5
& \tokup{9,583}{0.04\%} \\

\bottomrule
\end{tabular}
\vspace{-1.0em}
\end{wraptable}

\paragraph{TACO improves token efficiency across model scales.}
Tab.~\ref{tab:compression_efficiency} reports accuracy, average steps per run, and average prompt tokens per step on TB~2.0. For models with more than 200B parameters, such as Qwen3-Coder-480B and DeepSeek-V3.2, TACO reduces per-step token cost by approximately 10\% without substantially changing the average number of steps, indicating that high-capacity models can effectively leverage compressed context while maintaining stable reasoning trajectories. 
For models with fewer than 40B parameters, the per-step token reduction is relatively marginal. Since these models often fail prematurely on complex terminal tasks, TACO enables longer and more successful interaction trajectories by improving their use of terminal feedback, which naturally increases total steps and overall token consumption compared with their early-failure baselines.

\subsection{Efficiency Comparison}
\begin{wrapfigure}{r}{0.57\textwidth}
    \vspace{-4.1em}
    \centering
    \includegraphics[width=0.56\textwidth]{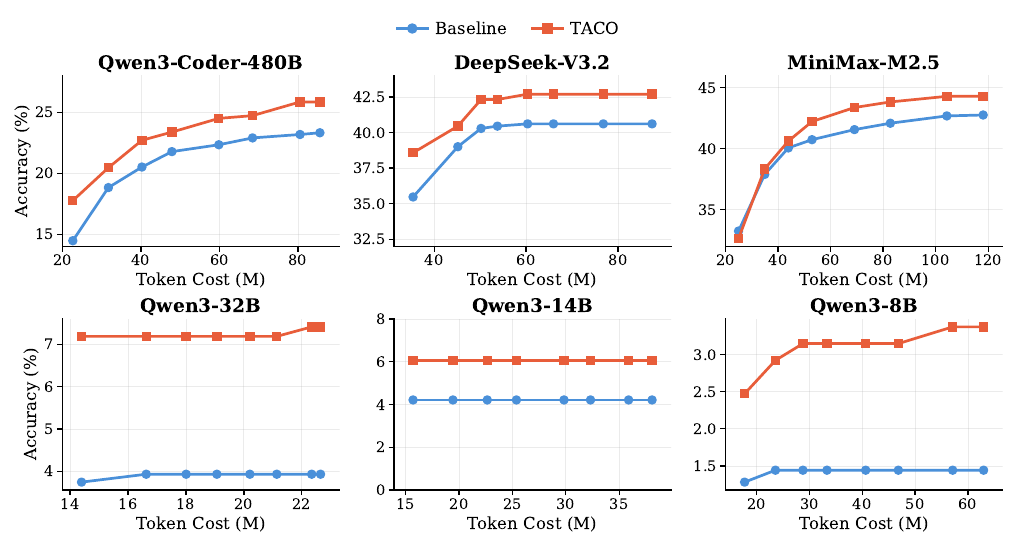}
    \caption{Agent Accuracy Under Identical Token Budgets.}
    \label{fig:acc_vs_token}
    \vspace{-2.45em}
\end{wrapfigure}

Considering differences in model capability, we set the budget range for each model according to the token-usage distribution observed under the vanilla agent scaffold (i.e., Terminus-2).

\paragraph{TACO improves accuracy under fixed token budgets.}
We evaluate the token efficiency of agents with and without \agentname{} on TB~2.0. As shown in Fig.~\ref{fig:acc_vs_token}, TACO consistently improves performance across all six models under the same token budgets, ranging from 14 million to 120 million tokens. For Qwen3-Coder-480B, DeepSeek-V3.2, and MiniMax-M2.5, TACO yields stable gains of 1\%--2\% across all budget settings, while for smaller models below 32B parameters, the gains are generally around 2\%--3\%.
We further investigate the impact of \agentname{} on model potential. As shown in Appendix~\ref{appdex:TTS}, \agentname{} substantially enhances the agent's test-time scaling capability.

\subsection{Comparison with Static Compression Methods}
\label{sec:static_baselines}

        \begin{wraptable}{r}{0.5\textwidth}
\vspace{-1.0em}
\centering
\caption{Static compression baselines on TB~2.0.
}
\label{tab:static_baselines}
\scriptsize
\setlength{\tabcolsep}{3.5pt}
\renewcommand{\arraystretch}{1.15}
\begin{tabular}{@{}lcc@{}}
\toprule
\textbf{Method} & \textbf{Acc.} & \textbf{Tok. Red.} \\
\midrule
Baseline  & 23.9 $\pm$ 2.9 & 0.00 \\
+ HQ Rule   & 24.3 $\pm$ 2.2 & 17.90 \\
+ LLM-Gen Rule & 19.71$\pm$ 1.6 & 7.10  \\
+ LLM Sum.  & 20.3 $\pm$ 3.2 & 21.30 \\
+ \agentname{} & \textbf{25.9 $\pm$ 1.5} & \textbf{10.78} \\
\bottomrule
\end{tabular}
\vspace{-1.0em}
\end{wraptable}
We compare \agentname{} with three static compression baselines on TB~2.0 using Qwen3-Coder-480B. 
These baselines cover representative static heuristic compression strategies for terminal observations, including command-aware truncation and error-preserving compression. 
Specifically, we consider \textbf{High-Quality Rules}, which use 200 human-curated structured rules; \textbf{LLM-Gen Rules}, which ask an LLM to generate task-conditioned rules but keep them fixed during execution; and \textbf{LLM Summarization}, which directly summarizes terminal observation with an LLM. 
The construction of the High-Quality Rules is described in Appendix~\ref{app:hq_static_rules}, and LLM-based compression prompts are provided in Appendix~\ref{appendix:prompt_static_baselines}.

As shown in Tab.~\ref{tab:static_baselines}, stronger token reduction does not necessarily lead to better task performance. 
Although LLM Summarization and the 200 human-curated rules remove more tokens than \agentname{}, they yield smaller accuracy gains, suggesting that terminal observation compression should prioritize preserving task-relevant signals over maximizing compression ratio. 
Moreover, both instruction-only LLM-generated rules and statically predefined rules show a substantial gap compared with \agentname{}, highlighting the importance of interactively refining compression rules based on environmental feedback and compression outcomes. 
By combining task-time adaptation with reusable rules from the Global Rule Pool, \agentname{} selectively removes low-value terminal noise while retaining information needed for later decisions.

\subsection{Convergence Metric Validation}
\label{Convergence Metric Validation}

\begin{wrapfigure}{r}{0.38\textwidth}
    \vspace{-5.2em}
    \centering
    \includegraphics[width=0.38\textwidth]{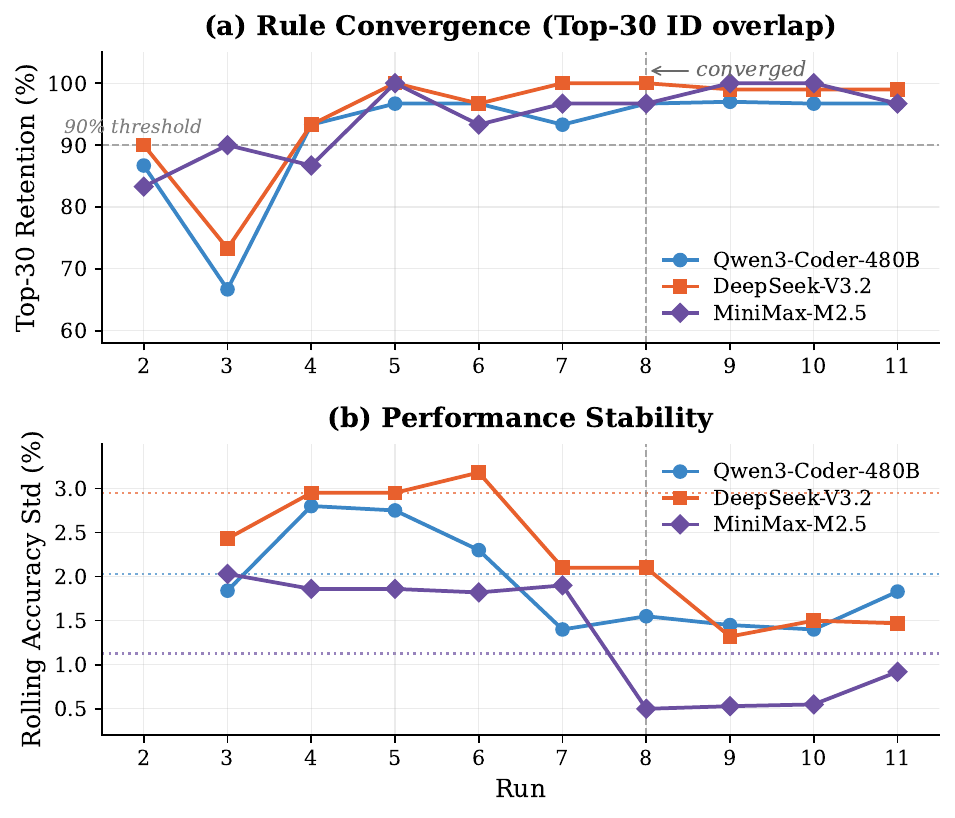}
    \caption{
    Rule-frontier convergence and performance stability.
    }
    \label{fig:convergence}
    \vspace{-1.0em}
\end{wrapfigure}

We propose a convergence metric, \emph{Retention}, to determine whether the self-evolution process has converged. To assess its effectiveness, we use the sliding standard deviation of task accuracy as an empirical measure of performance stability. The detailed computation procedure for the sliding standard deviation is provided in Appendix~\ref{metrics_details}.

As shown in Fig.~\ref{fig:convergence}, on TB~2.0, the sliding standard deviation decreases substantially once convergence is detected by \emph{Retention}. Across all three models, it drops from above 2.0 before convergence to around 1.0 afterward, indicating that \emph{Retention} provides a reliable signal of convergence in the self-evolution process.

\subsection{The Quality of Rule Generated by TACO}



\paragraph{Observation-level compression quality.}

\begin{wraptable}{r}{0.36\columnwidth}
\vspace{-1.8em}
\centering
\caption{Task-solving number changes.}
\label{tab:swap_analysis}
\scriptsize
\setlength{\tabcolsep}{5pt}
\renewcommand{\arraystretch}{0.95}
\begin{tabular}{lcc}
\toprule
\textbf{Model} & \textbf{Rescued} & \textbf{Regressed} \\
\midrule
Qwen3-8B          & \textbf{1} & 0 \\
Qwen3-14B         & \textbf{2} & 0 \\
Qwen3-32B         & \textbf{1} & 0 \\
Qwen3-Coder-480B  & \textbf{5} & 2 \\
DeepSeek-V3.2     & \textbf{8} & 4 \\
MiniMax-M2.5      & \textbf{4} & 3 \\
\bottomrule
\end{tabular}
\vspace{-1.6em}
\end{wraptable}

To assess whether \agentname{} preserves information needed for subsequent decisions, we randomly sample \(N=200\) observation pairs before and after compression from Qwen3-Coder-480B trajectories on TB~2.0.
For each pair, an LLM judge evaluates three aspects: preservation of potentially task-critical information, removal of redundancy, and retention of useful non-critical context.
The full protocol and judge prompt are provided in Appendix~\ref{app:audit}.
Evaluation results indicate that \agentname{} preserves potentially critical information in 96.0\% of cases, removes redundant content in 81.0\%, and retains useful non-critical context in 78.5\%. Only 4.0\% of cases are flagged as potential critical losses.
We manually inspect all eight flagged cases in Appendix~\ref{app:residual_loss}: seven are benign event-level flags with no observable behavioral impact, while the remaining case triggers the intended \texttt{need\_full\_output} complaint signal, providing feedback for subsequent rule evolution.
These results demonstrate the robustness of \agentname{} for preservation-aware context compression.

\paragraph{Task-level rescue and regression.}

We analyze task-solving number changes between Baseline and \agentname{} on TB~2.0. 
A task is counted as \emph{rescued} if Baseline fails but \agentname{} succeeds, and as \emph{regressed} if Baseline succeeds but \agentname{} fails. 
To reduce randomness, we average outcomes over the last five converged runs and consider a task solved only if at least 50\% of runs solve it. 
As shown in Tab.~\ref{tab:swap_analysis}, \agentname{} rescues more tasks than it regresses across all models, suggesting that adaptive observation compression more often improves execution than harms it by filtering redundant observations and preserving useful context for later decisions.



\subsection{Rule Reusability}

\begin{wraptable}{r}{0.35\textwidth}
\centering
\vspace{-5.7em} 
\caption{Rule reusability on TB~2.0.}
\label{tab:rule_re_use}
\scriptsize 
\setlength{\tabcolsep}{3pt}
\renewcommand{\arraystretch}{1.1}
\begin{tabular}{llc}
\toprule
\textbf{Model} & \textbf{Setting} & \textbf{Acc. (\%)} \\
\midrule
\multirow{3}{*}{DeepSeek-V3.2} 
& Baseline & 40.62 \\
& TACO (Online) & \textbf{42.77} \\
& TACO (Reuse) & 42.69 \\
\midrule
\multirow{3}{*}{Qwen3-480B} 
& Baseline & 23.90 \\
& TACO (Online) & 25.86 \\
& TACO (Reuse) & \textbf{26.96} \\
\midrule
\multirow{3}{*}{MiniMax-M2.5} 
& Baseline & 42.80 \\
& TACO (Online) & 44.16 \\
& TACO (Reuse) & \textbf{44.94} \\
\bottomrule
\end{tabular}
\vspace{-1.5em}
\label{Rule reusability}
\end{wraptable}

Although \agentname{} requires multiple cold-start iterations to reach rule convergence, this cost is incurred only once. 
Once converged, the learned Global Rule Pool can be frozen and reused for subsequent tasks, avoiding repeated test-time scaling. 
As shown in Tab.~\ref{Rule reusability}, the reused rules remain effective on TB~2.0, indicating that they capture generalizable terminal-output patterns rather than benchmark-specific artifacts. 
Therefore, cold-start inference cost can be amortized over future deployments.

\subsection{Ablation Study}


To investigate the contribution of self-evolution, we remove the two evolutionary components of TACO, namely \textbf{Intra-Task Rule Set Evolution} (ITRSE) and \textbf{Global Rule Pool Evolution} (GRPE), and evaluate the resulting variants on TB~2.0 with DeepSeek-V3.2.

For the variant without \textbf{Intra-Task Rule Set Evolution}, we use the final Global Rule Pool from TACO on TB~2.0 as a fixed initialization source for each task. During execution, the intra-task rule set remains unchanged, and neither task-specific rules nor the Global Rule Pool are updated. For the variant without \textbf{Global Rule Pool Evolution}, we perform only task-level rule evolution within each task, without maintaining or updating a shared Global Rule Pool.
\begin{wraptable}{r}{0.4\textwidth}
\vspace{-1.5em}
\centering
\caption{
Ablation results on TB~2.0. Total tokens are reported in millions (M).
}
\label{tab:ablation}
\scriptsize
\setlength{\tabcolsep}{4.0pt}
\renewcommand{\arraystretch}{1.05}
\begin{tabular}{@{}lccc@{}}
\toprule
\textbf{Method} & \textbf{Acc.} & \textbf{Total Tok. (M)} \\
\midrule
Baseline          & 40.6\%   & 99.60 \\
TACO w/o GRPE     & 40.4\%   & 81.57 \\
TACO w/o ITRSE    & 38.9\%   & \textbf{69.02} \\
\textbf{TACO (Full)} 
                  & \textbf{42.7\%}   & 81.45 \\
\bottomrule
\end{tabular}
\vspace{-0.8em}
\end{wraptable}
As shown in Tab.~\ref{tab:ablation}, using either component alone can reduce total token consumption, but it also leads to degraded task performance. This indicates that purely static compression is insufficient for terminal contexts, while rules derived from individual tasks alone have limited quality and generalizability. In contrast, TACO evolves a shared Global Rule Pool, enabling it to accumulate high-quality, reusable, and generalizable compression rules.


\section{Related Work}


\paragraph{Context Compression for Code and Terminal.}
As LLM agents move toward long-horizon software-engineering and terminal-based tasks~\citep{jimenez2024swebench,merrill2026terminal,yang2024swe,wang2024openhands,swwu2026terminalagentictrajectory}, context compression becomes increasingly important for reducing cost and improving reliability.
General prompt-compression methods, including token pruning, information filtering, retrieval-based selection, and LLM summarization~\citep{jiang2023llmlingua,pan2024llmlingua2,li2023selective}, shorten inputs by removing redundant content.
However, terminal observations pose a distinct safety challenge: verbose logs, compiler traces, test reports, and shell outputs interleave low-value noise with sparse but exact execution evidence, such as error messages, file paths, test names, command arguments, and package versions.
Generic pruning may discard these critical strings, while abstractive summarization may paraphrase or omit them.
Recent code-oriented compressors, such as LongCodeZip and SWE-Pruner~\citep{shi2025longcodezip,wang2026swe}, are closer to our motivation, but mainly target source-code or repository contexts and often rely on program structure or trained pruning models.

\paragraph{Agent Context Management and Self-Evolving Agents.}
Another related direction studies how long-horizon agents manage accumulated interaction histories.
Existing methods use trajectory summarization, memory retrieval, context truncation, observation masking, proactive context folding, or learned history compression to retain, summarize, fold, or remove past messages, actions, observations, and intermediate reasoning traces~\citep{liu2025context,kang2025ACON,ye2025agentfold,wan2025compass}.
These approaches help control growing histories, but they primarily operate after observations have already entered the agent context.
Meanwhile, training-free self-evolving agents improve future task solving by accumulating reusable memories, skills, plans, tools, or symbolic artifacts~\citep{zhou2026mementoskillsletagentsdesign,zhou2024symboliclearningenablesselfevolving}.

In contrast, \agentname{} compresses terminal observations before they enter the agent history and evolves reusable, preservation-aware compression rules, rather than task-solving strategies.

\section{Conclusion}
\label{sec: Conclusion}
We present \agentname{}, a plug-and-play, self-evolving terminal observation compression framework for terminal agents. By automatically discovering, refining, and reusing compression rules from interaction trajectories, \agentname{} enables adaptive and training-free context compression across diverse terminal environments. Experiments on TerminalBench and additional terminal-related benchmarks show that \agentname{} consistently improves both task performance and token efficiency across different agent frameworks and backbone models. These results highlight the importance of removing redundant terminal context for long-horizon reasoning and suggest a practical path toward more efficient and effective terminal agents. 



\bibliographystyle{plainnat}
\bibliography{references}

\clearpage

\appendix

\section{Limitations}
\label{limitation}
\agentname{} operates at the observation-compression layer rather than the model-capability layer. 
It does not modify the backbone model or improve its intrinsic capability; instead, it helps existing agents make better use of their context budget by presenting cleaner and less redundant terminal observations. 

\section{Ethics Statement}
\label{app:ethics}

Our experiments are conducted on publicly available software-engineering and terminal-agent benchmarks, and do not involve private user data or personally identifiable information. 
The proposed \agentname{} method itself does not rely on manual rule engineering. 
Human annotation was used solely to construct the High-Quality Rule static comparison baseline. 
Specifically, we recruited three Master's students in computer science to write structured compression rules based on sampled terminal-agent trajectories and predefined guidelines. 
This process involved only public benchmark trajectories and terminal outputs; it did not entail sensitive personal data, user profiling, or human-subject behavioral studies. 
The annotators were compensated at approximately \$19 per hour, which is above the applicable local minimum wage.

\section{Compute Resources}
\label{Compute resources}
\agentname{} is training-free and does not require additional model training.
All reported results are averaged over five independent runs unless otherwise
stated. All backbone models are served locally on a 20$\times$NVIDIA H800 GPU
cluster.

\section{Broader impacts}
\label{Broader impacts}
This work studies observational context compression for terminal agents and is not a standalone deployed system. Its positive impacts include reducing compute costs and improving the efficiency of long-horizon workflows. Any potential risks are strictly inherent to the underlying terminal agents rather than the compression mechanism itself. We therefore recommend deploying TACO alongside standard safety protocols, such as sandboxed execution environments and human oversight.

\section{Statistical Details}
\label{metrics_details}

We report the mean accuracy and run-level standard deviation over five evaluation runs.
The results are summarized in Tab.~\ref{tab:tb_std} and Tab.~\ref{tab:downstream_std}.

\begin{table}[h]
\centering
\caption{Accuracy (\%) and run-level standard deviation on TerminalBench.
All values are computed over five evaluation runs.}
\label{tab:tb_std}
\setlength{\tabcolsep}{6pt}
\small
\begin{tabular}{llcc}
\toprule
\multirow{2}{*}{Model} & \multirow{2}{*}{Method}
& TB~1.0 & TB~2.0 \\
\cmidrule(lr){3-3} \cmidrule(lr){4-4}
 & & Acc. (\%) & Acc. (\%) \\
\midrule

\multirow{2}{*}{MiniMax-M2.5}
& Baseline     & $42.30 \pm 3.94$ & $42.80 \pm 2.54$ \\
& \agentname{} & $45.25 \pm 1.76$ & $44.16 \pm 0.86$ \\

\multirow{2}{*}{DeepSeek-V3.2}
& Baseline     & $43.93 \pm 2.98$ & $40.62 \pm 2.88$ \\
& \agentname{} & $46.25 \pm 2.62$ & $42.77 \pm 2.54$ \\

\multirow{2}{*}{Qwen3-Coder-480B}
& Baseline     & $37.50 \pm 3.93$ & $23.90 \pm 1.48$ \\
& \agentname{} & $38.50 \pm 2.23$ & $25.86 \pm 0.78$ \\

\multirow{2}{*}{Qwen3-8B}
& Baseline     & $8.86 \pm 2.27$  & $1.43 \pm 1.31$ \\
& \agentname{} & $9.22 \pm 1.01 $  & $3.67 \pm 0.98$ \\

\multirow{2}{*}{Qwen3-14B}
& Baseline     & $5.23 \pm 2.15$  & $4.04 \pm 2.23$ \\
& \agentname{} & $11.25 \pm 2.32$ & $6.15 \pm 1.67$ \\

\multirow{2}{*}{Qwen3-32B}
& Baseline     & $11.25 \pm 2.29$ & $3.92 \pm 3.13$ \\
& \agentname{} & $14.13 \pm 1.79$ & $7.48 \pm 2.44$ \\

\bottomrule
\end{tabular}
\end{table}

\begin{table}[t]
\centering
\caption{Accuracy (\%) and run-level standard deviation on downstream
terminal-related benchmarks. All values are computed over five evaluation runs.}
\label{tab:downstream_std}
\setlength{\tabcolsep}{6pt}
\small
\begin{tabular}{llc}
\toprule
\multirow{2}{*}{Benchmark} & \multirow{2}{*}{Method}
& MiniMax-M2.5 \\
\cmidrule(lr){3-3}
 & & Acc. (\%) \\
\midrule

\multirow{2}{*}{SWE-Bench Lite}
& Baseline     & $56.30 \pm 2.32$ \\
& \agentname{} & $57.12 \pm 2.19$ \\

\multirow{2}{*}{CompileBench}
& Baseline     & $75.00 \pm 8.69$ \\
& \agentname{} & $75.00 \pm 7.82$ \\

\multirow{2}{*}{DevEval}
& Baseline     & $38.10 \pm 1.56$ \\
& \agentname{} & $39.74 \pm 1.17$ \\

\multirow{2}{*}{CRUST-Bench}
& Baseline     & $47.00 \pm 3.43$ \\
& \agentname{} & $48.05 \pm 2.78$ \\

\bottomrule
\end{tabular}
\end{table}

\section{Additional TerminalBench Results with Closed-Source Models}
\label{app:closed_source_tb}

For reference, we report closed-source model results on TerminalBench in Tab.~\ref{tab:closed_source_tb}. 

\begin{table}[h]
\centering
\caption{Closed-source model results on TerminalBench. \textsuperscript{*} indicates results reported on the TerminalBench leaderboard.}
\label{tab:closed_source_tb}
\footnotesize
\setlength{\tabcolsep}{6pt}
\renewcommand{\arraystretch}{1.05}
\begin{tabular}{lccc}
\toprule
\textbf{Model} & \textbf{Agent Scaffold} & \textbf{TB~1.0} & \textbf{TB~2.0} \\
\midrule
Gemini-3-Flash-Preview & Terminus-2 & 53.72 & 51.70\textsuperscript{*} \\
Gemini-3-Pro-Preview   & Terminus-2 & 46.35 & 56.90\textsuperscript{*} \\
Claude-Opus-4.5        & Terminus-2 & 47.50 & 57.80\textsuperscript{*} \\
Claude-Sonnet-4.5      & Terminus-2 & 51.00\textsuperscript{*} & 42.80\textsuperscript{*} \\
GPT-5.1                & Terminus-2 & 35.50 & 47.60\textsuperscript{*} \\
GPT-5.2                & Terminus-2 & 54.38 & 54.00\textsuperscript{*} \\
\bottomrule
\end{tabular}
\end{table}

\section{Hyperparameter Selection}
\label{subsec:Hyperparameter Selection}

\begin{figure*}[h]
    \centering
\includegraphics[width=0.89\textwidth,height=0.62\textheight,keepaspectratio]{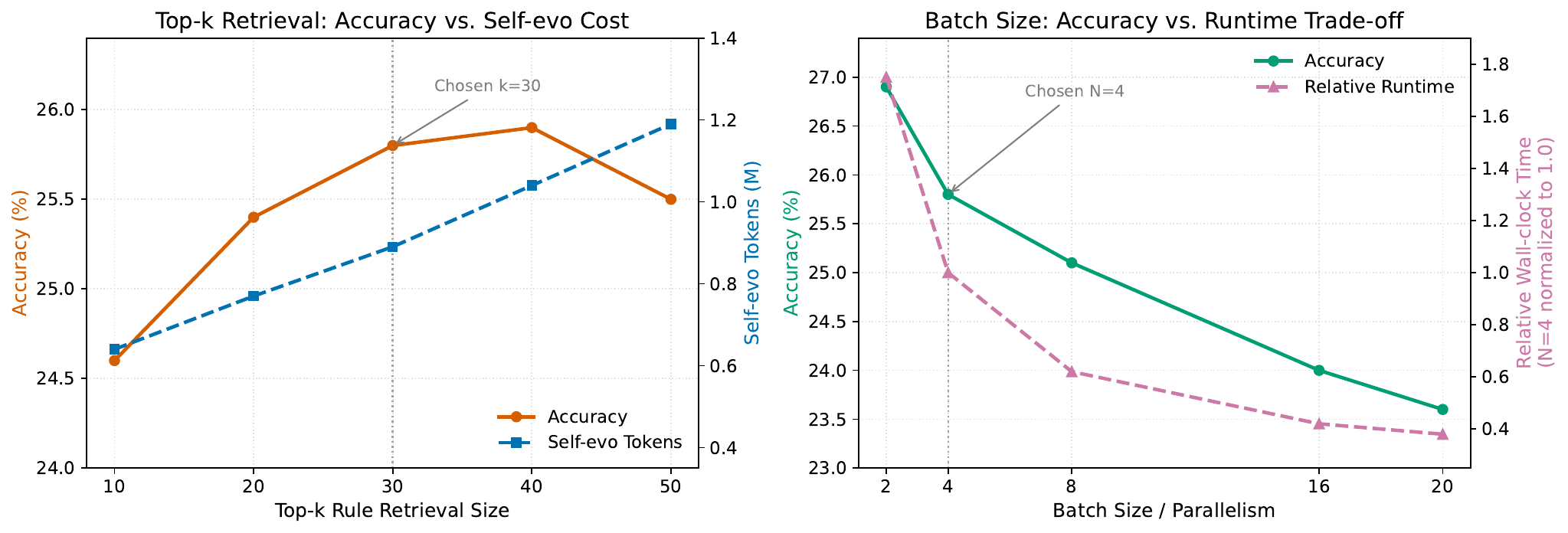}
    \caption{Hyperparameter selection for TACO. Left: effect of the top-$k$ rule retrieval size on accuracy and self-evolution token cost. Right: effect of batch size $N$ on accuracy and relative runtime. The self-evolution token cost and relative runtime shown in the figure are reported as relative values, normalized to the results with $k=30$ and $N=4$.}
    \label{fig:hyperparameter_selection}
\end{figure*}

The batch size \textbf{$N$} in TACO’s parallel execution and the \textbf{Top-$k$} hyperparameter for task-specific rule initialization both shape the interaction between the Intra-Task Rule Pool and the Global Rule Pool, and thus affect the final performance. To examine their effects, we perform an ablation study on TB~2.0 with Qwen3-Coder-480B.

\paragraph{Top-$k$ Selection.}
To examine the effect of Top-$k$ under a fixed computational budget, we fix the batch size at $N=4$ and vary Top-$k$. We evaluate both agent accuracy and the token cost of rule selection. As shown in the left part of Fig.~\ref{fig:hyperparameter_selection}, a smaller $k$ reduces token consumption but limits performance by narrowing the pool of candidate historical rules. As $k$ increases, accuracy improves at first and then gradually saturates, while the cost of self-evolution continues to rise. When $k>30$, the additional performance gain becomes marginal or even negative, whereas the token cost keeps increasing. Moreover, a larger $k$ leads to a longer retrieval context, which may be less stable for smaller models. Therefore, we set $k=30$ in the main experiments.

\paragraph{Batch Size $N$ Selection.}
We fix Top-$k$ at 30 and investigate the effect of different batch sizes $N$ on TB~2.0. As shown in the right part of Fig.~\ref{fig:hyperparameter_selection}, smaller batch sizes tend to yield higher accuracy, since newly learned rules can be written back more frequently and reused by subsequent tasks sooner. In contrast, larger batch sizes improve parallel throughput and overall runtime efficiency, but delay the application of newly acquired rules to later tasks. Balancing accuracy, throughput, and rule-propagation efficiency, we adopt $N=4$ in the main experiments.

\section{Best-of-K Trajectory Sampling}
\label{appdex:TTS}
\begin{figure*}[!t]
    \centering
    \includegraphics[width=0.85\textwidth]{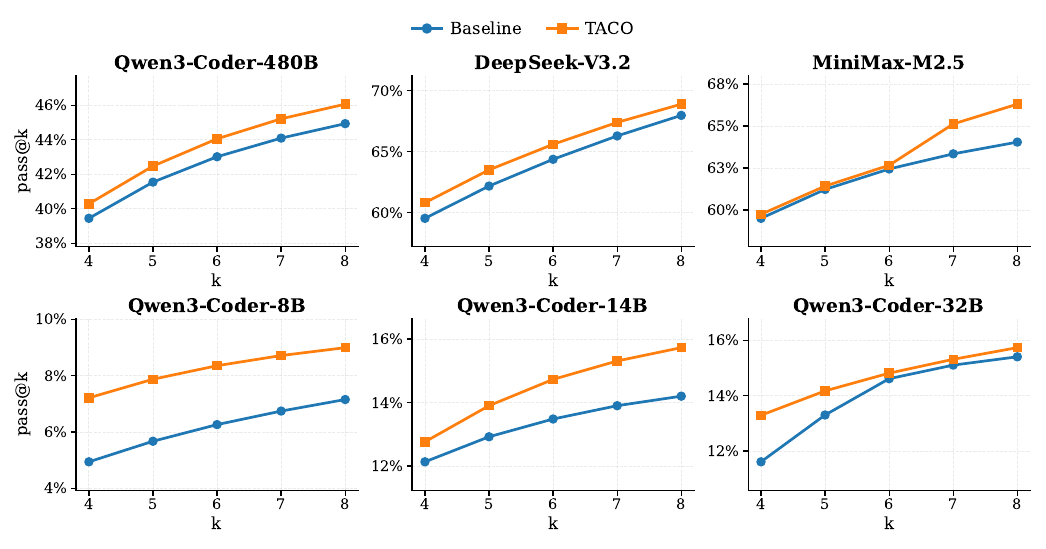}
    \caption{
    Pass@k comparison between Baseline and \agentname{} across six models on TB~2.0.
    \agentname{} consistently improves pass@k across all evaluated settings.
    }
    \label{fig:pass_at_k_comparison}
    \vspace{-0.6em}
\end{figure*}

Test-time scaling (e.g., Best-of-K) increases inference-time compute by sampling multiple candidate solutions, often leading to substantial performance gains. 
Pass@k, a Best-of-K-style metric, has been widely used to evaluate agents' potential ability~\citep{wu2024comparativestudyreasoningpatterns,zhu2025scalingtesttimecomputellm,snell2024scaling}.
Therefore, to further demonstrate that \agentname{} improves agents' potential ability, we compare pass@k for $k \in \{4,5,6,7,8\}$ across six models on TB~2.0, with and without \agentname{}.
As shown in Fig.~\ref{fig:pass_at_k_comparison}, TACO consistently improves pass@k across all models and k, suggesting that its compression rules improve the quality of individual trajectories and remain complementary to test-time scaling.

\section{Information Used in Reward-Free Unsupervised Rule Evolution}
\label{appdix:Leakage}

\begin{table}[t]
\centering
\small
\begin{tabular}{lc}
\toprule
\textbf{Information Source} & \textbf{Used by TACO} \\
\midrule
Terminal observation & Yes \\
Executed command & Yes \\
Rule application count & Yes \\
\midrule
Final task success/failure & No \\
Hidden tests or verifier result & No \\
Ground-truth answer or patch & No \\
Leaderboard score & No \\
\bottomrule
\end{tabular}
\caption{Information sources used during TACO rule evolution. TACO updates compression rules only from observation-level interaction signals and does not access benchmark correctness signals.}
\label{tab:leakage_protocol}
\end{table}

As shown in Tab.~\ref{tab:leakage_protocol}, we present the information used for rule evolution.
This table formalizes the boundary of TACO’s unsupervised adaptation: it may use observable interaction traces, but not correctness, reward, or answer signals.



\section{Token Overhead of Rule Evolution}
\label{app:token_overhead}

To ensure fair cost accounting, all token numbers reported in the main paper include both backbone-agent execution tokens and auxiliary LLM calls introduced by \agentname{} for rule selection, generation, and revision.
As shown in Tab.~\ref{tab:token_overhead}, the auxiliary rule-evolution overhead remains below 2\% across all benchmarks, indicating that the reported token reductions are net end-to-end savings rather than excluding the cost of self-evolution.

\begin{table}[t]
\centering
\caption{
Auxiliary token overhead of \agentname{} rule evolution.
All main-paper token counts include this overhead.
}
\label{tab:token_overhead}
\footnotesize
\setlength{\tabcolsep}{6pt}
\begin{tabular}{lc}
\toprule
\textbf{Benchmark} & \textbf{Aux. Tok. / Total Tok.} \\
\midrule
SWE-Bench Lite & 1.07\% \\
CompileBench   & 1.19\% \\
CRUST-Bench    & 0.61\% \\
DevEval        & 1.88\% \\
TB~1.0         & 0.94\% \\
TB~2.0         & 0.82\% \\
\bottomrule
\end{tabular}
\end{table}

\section{Case Study}
\label{case_study}

To provide a concrete illustration of TACO's self-evolving compression behavior, we examine three successful trajectories from TerminalBench~2.0, executed by the Terminus-2 agent with Qwen3-Coder-480B as the backbone. The selected tasks---\texttt{adaptive-rejection-sampler} (statistical computing in R), \texttt{sqlite-with-gcov} (build-system instrumentation), and \texttt{vulnerable-secret} (binary reverse engineering)---represent distinct terminal domains with qualitatively different output patterns. All three achieve a reward of 1.0, while TACO's compression mechanism reduces substantial token overhead throughout. We analyze these trajectories from three perspectives: (1) how task-specific rules are initialized and evolved online, (2) the quantitative compression statistics observed in the logs, and (3) concrete before-and-after comparisons demonstrating the transition from syntactic truncation to semantic filtering.

\subsection{Task-Specific Rule Initialization and Online Evolution}

As described in Sec.~4.2, TACO initializes the task-specific active rule set $R_t$ by retrieving top-ranked rules from the Global Rule Pool $R_g$ and invoking an LLM to select, adapt, and generate new rules conditioned on the task description. Tab.~\ref{tab:case_rule_init} summarizes the rule composition for each trajectory.

\begin{table}[h]
\centering
\caption{Rule composition across three trajectories. ``New (plan-time)'' denotes task-specific rules generated at initialization; ``New (mid-task)'' denotes rules evolved reactively via the Intra-Task Rule Set Evolution mechanism (Sec.~4.4).}
\label{tab:case_rule_init}
\small
\begin{tabular}{lccc}
\toprule
\textbf{Task} & \textbf{New (plan-time)} & \textbf{New (mid-task)} \\
\midrule
\texttt{adaptive-rejection-sampler}   & 2 & 1 \\
\texttt{sqlite-with-gcov}   & 2 & 0 \\
\texttt{vulnerable-secret}  & 2 & 1 \\
\bottomrule
\end{tabular}
\end{table}

\textbf{Plan-time rule generation.} The LLM generates task-specific rules that embed domain understanding directly into the compression logic, going beyond syntactic heuristics. For \texttt{sqlite-with-gcov}, the rule \texttt{sqlite\_make\_gcov\_mod} targets \texttt{make} output with coverage flags---it strips per-object \texttt{gcc} invocation lines and \texttt{cp} file-listing commands, while explicitly retaining gcov-specific errors (\texttt{cannot find -lgcov}, \texttt{profiling error}) and the final linking step. For \texttt{vulnerable-secret}, the rule \texttt{secret\_key\_extractor} aggressively filters all output from \texttt{/app/} executables except lines matching \texttt{FLAG\{.*\}}, tracebacks, segmentation faults, and error signals---effectively transforming the compressor into a targeted information extractor for the hidden secret. For \texttt{adaptive-rejection-sampler}, the rule \texttt{pytest\_stochastic\_testing\_mod} strips verbose \texttt{PASSED} lines and session headers from test output while preserving statistical validation signals (KS test p-values, chi-square results, sample mean/variance).

\textbf{Reactive mid-task evolution.} In two trajectories, TACO's Intra-Task Rule Set Evolution (Sec.~4.4) triggered reactive rule generation when the agent encountered lengthy terminal observation that were not covered by any active rule. 
In \texttt{adaptive-rejection-sampler}, the initial \texttt{apt\_install} rule only partially compressed the output of \texttt{apt-get install -y r-base}, reducing the first chunk from 10,071 to 5,371~characters (ratio 0.53). 
However, a later continuation chunk remained largely uncompressed, despite being dominated by repetitive ``\texttt{Unpacking}'' and ``\texttt{Setting up}'' lines for over 200 packages. 
This exposed a coverage gap in the active rule rather than a need to preserve the full output, prompting TACO to refine the rule to collapse repeated package-lifecycle logs while retaining final status and error signals.
TACO evolved \texttt{apt\_install\_unpacked\_packages}, which matched and suppressed these repetitive lines, reducing the 10,071-character continuation to 73~characters (ratio 0.007)---a 99.3\% reduction. In \texttt{vulnerable-secret}, the first \texttt{objdump -d} call at episode~9 produced 5,169~characters of disassembly with no matching rule. TACO generated \texttt{objdump\_disassembly\_rule}, which strips repetitive hex-dump instruction lines while preserving section headers, symbol labels (e.g., \texttt{<gets@plt>}), and address markers. This rule then fired on 18 subsequent \texttt{objdump} invocations across the remaining 92~episodes.

The \texttt{sqlite-with-gcov} task did not require new rules during execution. 
The rules initialized at the beginning of the task were sufficient to handle its terminal outputs, including compiler outputs and git-related noise. 
This indicates that \agentname{} only needs online rule evolution when it encounters output patterns not covered by the initialized rule set.

\subsection{Quantitative Compression Analysis}

Tab.~\ref{tab:case_stats} reports the compression statistics extracted from each trajectory.

\begin{table}[h]
\centering
\caption{Compression statistics for the three case study trajectories. ``Entries'' counts the number of steps where rule-based compression was applied. ``Chars saved'' denotes the total reduction in terminal observation characters. ``Best ratio'' is the minimum compression ratio observed (lower = more aggressive compression).}
\label{tab:case_stats}
\small
\begin{tabular}{lcccccc}
\toprule
\textbf{Task} & \textbf{Episodes} & \textbf{Entries} & \textbf{Chars saved} & \textbf{Overall ratio} & \textbf{Best ratio} \\
\midrule
\texttt{adaptive-rejection-sampler} & 25 & 4 & 14,766 & 0.285 & 0.007 \\
\texttt{sqlite-with-gcov} & 22 & 9 & 9,647 & 0.464 & 0.390 \\
\texttt{vulnerable-secret} & 101 & 25 & 29,636 & 0.455 & 0.146 \\
\bottomrule
\end{tabular}
\end{table}

Several observations stand out. First, the compression behavior is highly non-uniform: in \texttt{adaptive-rejection-sampler}, only 4 out of 25~episodes triggered compression, but those 4~entries accounted for over 14,000~characters of savings---concentrated entirely on the high-volume \texttt{apt-get} installation steps. Second, in \texttt{vulnerable-secret}, the evolved \texttt{objdump\_disassembly\_rule} accounts for 18 of the 25~compression entries and 29,464 of the 29,636~characters saved, demonstrating how a single reactively evolved rule can dominate the compression profile of a long-horizon task. Across all 18~\texttt{objdump} applications, the average compression ratio is 0.445, with individual ratios ranging from 0.146 to 0.999 depending on the amount of disassembly output in each call.

\subsection{Before-and-After Comparison}

To illustrate how TACO transitions from na\"ive truncation to task-aware semantic filtering, we present three representative compression examples drawn directly from the trajectory logs.

\subsubsection{Package Installation (\texttt{apt-get install})}
In \texttt{adaptive-rejection-sampler} (episode~4), the agent installs the R runtime via \texttt{apt-get install -y r-base}, which triggers the download and configuration of 200+ dependency packages. The raw terminal observation is dominated by mechanically repetitive lines with no task-relevant content.

\textbf{Raw terminal observation} (10,071~characters):
\begin{lstlisting}[language=bash, basicstyle=\ttfamily\scriptsize, frame=single, breaklines=true]
Unpacking libc6-dev:amd64 (2.39-0ubuntu8.7) ...
Unpacking gcc-13-base:amd64 (13.3.0-6ubuntu2~24.04.1) ...
Unpacking libisl23:amd64 (0.26-3build1.1) ...
Unpacking libmpfr6:amd64 (4.2.1-1build1.1) ...
Unpacking libmpc3:amd64 (1.3.1-1build1.1) ...
Unpacking cpp-13-x86-64-linux-gnu (13.3.0-6ubuntu2~24.04.1) ...
[... 150+ Unpacking/Setting up lines ...]
Setting up r-base-core (4.3.3-2build2) ...
Setting up r-base (4.3.3-2build2) ...
Processing triggers for libc-bin (2.39-0ubuntu8.7) ...
\end{lstlisting}

\textbf{Compressed output} (73~characters, ratio = 0.007):
\begin{lstlisting}[language=bash, basicstyle=\ttfamily\scriptsize, frame=single, breaklines=true]
[WAITING] apt-get install -y r-base

Current status: Setting up x11-utils
\end{lstlisting}

The evolved rule \texttt{apt\_install\_unpacked\_packages} recognizes that the hundreds of ``\texttt{Unpacking}'' and ``\texttt{Setting up}'' lines carry no information needed for subsequent reasoning---the only signal the agent requires is whether the installation succeeded or is still in progress. By compressing the output to a 73-character status indicator, TACO prevents a 10,000-character log from displacing task-relevant context (the R code and test outputs the agent needs for upcoming steps). This illustrates the efficiency gain described in Sec.~6.1: observation compression reduces contextual redundancy, allowing the model to more easily identify the critical information needed for task completion.

\subsubsection{Build System Output (\texttt{make})}
In \texttt{sqlite-with-gcov} (episode~14), the agent executes \texttt{make} to compile SQLite with gcov coverage instrumentation. The raw output contains verbose file-copying commands alongside semantically important build configuration messages.

\textbf{Raw terminal observation} (6,519~characters):
\begin{lstlisting}[language=bash, basicstyle=\ttfamily\scriptsize, frame=single, breaklines=true]
$ make
touch .main.mk.checks
cc -g -o mksourceid .../tool/mksourceid.c
cc -g -o jimsh -O1 -DHAVE_REALPATH .../autosetup/jimsh0.c
./jimsh .../tool/mksqlite3h.tcl ... -o sqlite3.h
./jimsh .../tool/mkctimec.tcl
Overwriting ctime.c...
cc -g -o mkkeywordhash ... .../tool/mkkeywordhash.c
./mkkeywordhash > keywordhash.h
cc -g -o lemon .../tool/lemon.c
cp .../src/parse.y .
./lemon -DSQLITE_ENABLE_MATH_FUNCTIONS ... -S parse.y
cp -f .../src/alter.c .../src/analyze.c .../src/attach.c
  .../src/auth.c .../src/backup.c .../src/bitvec.c
  [... 100+ source files copied to tsrc/ ...]
rm -f tsrc/sqlite.h.in tsrc/parse.y
./jimsh .../tool/vdbe-compress.tcl <tsrc/vdbe.c >vdbe.new
touch .target_source
cc -fPIC -O2 -g -DSQLITE_COVERAGE_TEST=1 -fprofile-arcs
  -ftest-coverage -c sqlite3.c
\end{lstlisting}

\textbf{Compressed output} (3,183~characters, ratio = 0.488):
\begin{lstlisting}[language=bash, basicstyle=\ttfamily\scriptsize, frame=single, breaklines=true]
$ make
touch .main.mk.checks
cc -g -o mksourceid .../tool/mksourceid.c
[compiler output compressed -- long command lines truncated]
  [25 lines removed]
cc -g -o jimsh -O1 -DHAVE_REALPATH .../autosetup/jimsh0.c
./jimsh .../tool/mksqlite3h.tcl ... -o sqlite3.h
./jimsh .../tool/mkctimec.tcl
Overwriting ctime.c...
./jimsh .../tool/mkpragmatab.tcl
Overwriting pragma.h with new pragma table...
cc -g -o mkkeywordhash ... .../tool/mkkeywordhash.c
...
touch .target_source
cc -fPIC -O2 -g -DSQLITE_COVERAGE_TEST=1 -fprofile-arcs
  -ftest-coverage -c sqlite3.c
\end{lstlisting}

The evolved compiler-output rule removes the verbose \texttt{cp -f} file-listing block, which contains 25~lines enumerating over 100 source files copied to \texttt{tsrc/}, while preserving the semantically critical elements: tool-generation commands, build configuration messages such as ``\texttt{Overwriting ctime.c...}'', and---crucially---the final compilation command containing \texttt{-fprofile-arcs -ftest-coverage}. 
This final command confirms that gcov coverage instrumentation is correctly enabled. 
A positional truncation strategy would risk cutting from the end and losing this critical line, whereas \agentname{} retains it through rule-guided filtering regardless of its position.

\subsubsection{Binary Disassembly (\texttt{objdump})}
In \texttt{vulnerable-secret} (episode~35), the agent disassembles the target binary to trace control flow and locate the flag-extraction logic. The \texttt{objdump\_disassembly\_rule}---which was reactively generated at episode~9---is applied to compress the assembly output.

\textbf{Raw terminal observation} (5,619~characters):
\begin{lstlisting}[language=bash, basicstyle=\ttfamily\scriptsize, frame=single, breaklines=true]
$ objdump -d vulnerable | sed -n '150,250p'
  4011e5:  31 f6            xor    %esi,%esi
  4011e7:  bf 11 00 00 00   mov    $0x11,%edi
  4011ec:  31 c0            xor    %eax,%eax
  4011ee:  e8 7d fe ff ff   call   401070 <signal@plt>
  4011f3:  31 c0            xor    %eax,%eax
  4011f5:  e8 96 fe ff ff   call   401090 <ptrace@plt>
  4011fa:  48 85 c0         test   %rax,%rax
  4011fd:  0f 88 9d 00 00 00 js  4012a0 <exit@plt+0x220>
  [... 90+ more instruction lines with hex bytes ...]
\end{lstlisting}

\textbf{Compressed output} (821~characters, ratio = 0.146):
\begin{lstlisting}[language=bash, basicstyle=\ttfamily\scriptsize, frame=single, breaklines=true]
$ objdump -d vulnerable | sed -n '150,250p'
  4011e5:  31 f6            xor    %esi,%esi
  4011e7:  bf 11 00 00 00   mov    $0x11,%edi
  4011ec:  31 c0            xor    %eax,%eax
  4011ee:  e8 7d fe ff ff   call   401070 <signal@plt>
  4011f5:  e8 96 fe ff ff   call   401090 <ptrace@plt>
\end{lstlisting}

The rule achieves an 85.4\% reduction by stripping repetitive hex-dump lines that carry no symbolic significance, while preserving \texttt{call} instructions (which reveal API usage such as \texttt{signal} and \texttt{ptrace}), branch targets with symbol labels, and section headers. This is precisely the information the agent needs to trace the binary's anti-debugging logic and locate the flag-extraction code path. Across the full 101-episode trajectory, this single evolved rule saved 29,464~characters of redundant disassembly output---approximately 54\% of all compression savings in this task. For a trajectory that consumed 2.4M input tokens, preventing this volume of low-value content from entering the context window directly contributes to the agent's ability to maintain coherent long-horizon reasoning.

These trajectories jointly show that TACO acts as a domain-adaptive semantic compressor rather than a generic truncation strategy. By initializing reusable rules from the Global Rule Pool and evolving new rules online for uncovered outputs, TACO preserves task-critical signals while filtering large volumes of repetitive terminal noise. This provides a concrete explanation for the performance gains observed in the main experiments.

\section{Rule Format and Examples}
\label{Rule Format and Examples}

\subsection{Rule Schema and Evolved Examples}
\label{app:rule_schema}

Listing~\ref{lst:evolved_rules} presents a high-quality, self-evolved rule for \texttt{7z} archive extraction to illustrate the structural composition and semantic filtering capabilities of TACO. This autonomously generated rule captures complex domain-specific patterns, achieving maximum global confidence ($c_r^g = 1.0$) and high application counts without triggering any over-compression complaints.

\begin{lstlisting}[float=htbp, basicstyle=\ttfamily\tiny, frame=single, breaklines=true, caption={Examples of high-quality, self-evolved compression rules autonomously discovered by TACO. These rules demonstrate fine-grained, domain-specific semantic filtering capabilities.}, label={lst:evolved_rules}]
[
  {
    "rule_id": "seven_zip_extraction",
    "trigger_regex": "\\b7z\\b.*\\s+secrets\\.7z\\b",
    "description": "Compresses 7zip extraction output, preserving errors and success indicators while removing verbose file listing and progress information.",
    "keep_patterns": [
      "\\bError:",
      "\\bERROR:",
      "\\berror:",
      "Wrong password",
      "Can not open the file as",
      "Everything is Ok",
      "Extracting\\s+.*secret_file\\.txt",
      "Data Error in encrypted file"
    ],
    "strip_patterns": [
      "^\\s*[0-9]+ files?,",
      "^\\s*[0-9]+ folders?,",
      "^\\s*Size:\\s+",
      "^\\s*Compressed:\\s+",
      "^\\s*Processing archive:",
      "^\\s*--",
      "^\\s*$"
    ],
    "keep_first_n": 5,
    "keep_last_n": 5,
    "max_lines": null,
    "summary_header": "[7z extraction output compressed]",
    "priority": 42,
    "confidence": 1.0,
    "times_applied": 126,
    "times_complained": 0
  },
  {
    "rule_id": "nginx_setup_mod",
    "trigger_regex": "nginx|openssl|a2ensite|systemctl\\s+(start|enable|restart)\\s+nginx",
    "description": "Compresses Nginx and OpenSSL setup output, preserving errors, SSL generation notices, and service start/enable status",
    "keep_patterns": [
      "\\berror:",
      "\\bError:",
      "\\bERROR:",
      "Failed",
      "failed",
      "Traceback",
      "Generating.*cert",
      "server started",
      "server enabled",
      "service failed",
      "Job for nginx.service failed"
    ],
    "strip_patterns": [
      "Generating RSA private key",
      "writing new private key to",
      "-----BEGIN PRIVATE KEY-----",
      "-----END PRIVATE KEY-----",
      "Signature ok",
      "subject=",
      "Getting Private key",
      "\\s*\\*\\s+[^a-z].*"
    ],
    "keep_first_n": 5,
    "keep_last_n": 10,
    "max_lines": 30,
    "summary_header": "[Nginx/SSL setup output compressed]",
    "priority": 42,
    "confidence": 1.0,
    "times_applied": 122,
    "times_complained": 0
  }
]
\end{lstlisting}

\clearpage

\subsection{Construction of High-Quality Static Rules}
\label{app:hq_static_rules}

The High-Quality Rule method is designed as a strong static compression comparison. 
We construct 200 structured rules from multiple complete TB~2.0 trajectories using LLM-assisted drafting followed by manual review. 
These rules cover common terminal-output patterns such as installation logs, compilation traces, test outputs, progress bars, directory listings, and long file views. 
During evaluation, the rules are fixed and do not adapt to the current task, agent behavior, or over-compression feedback. 
This makes them a non-trivial static comparison while preserving the key distinction from \agentname{}'s online rule selection and self-evolution.

\section{Compression-Quality Evaluation Protocol}
\label{app:audit}

\subsection{Evaluation Protocol}
\label{app:audit_protocol}

We collect all compression events from \agentname{} runs with
Qwen3-Coder-480B on TerminalBench~2.0 and draw a uniform random
sample of 200 events. Each event corresponds to one raw terminal
observation and the compressed observation actually provided to the
agent. For each sampled event, we prompt a GPT-5.5-based judge with
the task instruction, the issued shell command, the original
uncompressed terminal observation, and the compressed output seen by the
agent.

The judge scores three independent binary properties:
\texttt{kept\_critical}, which corresponds to whether potentially task-critical information is preserved;
\texttt{removed\_useless}, which corresponds to whether redundant or noisy content is removed;
and \texttt{kept\_useful}, which corresponds to whether useful non-critical context is retained.
In Tab.~\ref{tab:compression_quality}, these are reported as ``Critical info preserved'', ``Redundant/noisy removed'', and ``Useful context retained'', respectively.
We additionally report ``Potential critical loss'' as the complement of \texttt{kept\_critical}.
The full prompt is shown below.

\begin{promptbox}{COMPRESSION-QUALITY JUDGE PROMPT}
[SYSTEM]
You are an expert reviewer for terminal-output compression in
coding agents.

You will be given ONE compression OBSERVATION:
  - the TASK INSTRUCTION the agent is solving
  - the COMMAND that produced the terminal observation
  - the ORIGINAL terminal observation (before compression)
  - the COMPRESSED terminal observation (what the agent actually saw)

Judge THREE INDEPENDENT binary properties for this observation:

(1) kept_critical  - Was critical info preserved in the compressed view?
    Critical = anything whose loss would BREAK the agent's ability to
    solve the task. e.g. error messages / stack traces, final test
    PASS/FAIL line, the actual numerical result asked for, the line
    the agent is grepping for, exit codes, the *body* of a newly
    written script / config file, the port number / URL / file path
    that's the deliverable, version strings.
    1 = critical info still readable in compressed view; 0 = lost.
    If there is no critical content in the original, default to 1.
    Note: if the missing content is something the agent itself just
    produced in its preceding message (e.g. the body of a
    `cat > f.py << EOF ... EOF` heredoc that the terminal merely
    echoes back), the agent already has it and you should score 1.

(2) removed_useless - Was noisy, useless info removed?
    Useless = bytes that carry no signal for the agent. e.g. ANSI
    escape codes, trailing empty newlines, apt's per-package
    "Setting up / Unpacking / Get:" middle lines, pip "Downloading
    ... 0% / 50% / 100%" progress bars, repeated bash empty-prompt
    polling lines, MOTD / login banner, duplicated headers.
    1 = a noticeable amount of noise was removed (or there was
        little noise to begin with AND compression didn't add noise);
    0 = compressed output is still mostly noise OR compression
        deleted real content instead of noise.

(3) kept_useful - Was useful (non-critical) context preserved?
    Useful = nice-to-have context that helps the agent reason
    without re-running a command. e.g. echoed full command line,
    key warnings, a few representative lines of a large list,
    dependency names, directory tree summary, first/last lines of
    a long log.
    1 = enough useful context kept that the agent can proceed
        without re-querying;
    0 = compression so harsh the agent likely needs to re-issue
        commands.

Critical vs Useful:
  - Critical = task FAILS or wrong answer if dropped.
  - Useful   = task still solvable, but agent wastes turns or risks
               misinterpretation.

Be strict but fair. Each dimension is judged on its own.

[USER]
# Task instruction
{instruction}

# Command
`{command}`

# Original terminal observation  ({orig_len} chars)
{original}

# Compressed terminal observation  ({comp_len} chars, saved {saved})
{compressed}


# Output (one JSON object only, no extra prose, no markdown fence)
{"kept_critical":   0 or 1,
 "removed_useless": 0 or 1,
 "kept_useful":     0 or 1,
 "note": "<<=20 words explaining the judgment>"}
\end{promptbox}

\subsection{Residual Loss Analysis}
\label{app:residual_loss}
\begin{table}[h]
\centering
\caption{Manual inspection of the eight potential critical-loss events.}
\label{tab:residual_loss_cases}
\scriptsize
\setlength{\tabcolsep}{4pt}
\renewcommand{\arraystretch}{0.95}
\begin{tabular}{lcc}
\toprule
\textbf{Pattern} & \textbf{Count} & \textbf{Effect} \\
\midrule
Installation log confirmation folded & 5 & No visible behavior change \\
Long file inspection summarized & 2 & One triggers complaint \\
Heredoc echo omitted & 1 & Content already in history \\
\bottomrule
\end{tabular}
\end{table}

We manually inspect all eight events flagged as potential critical losses.
Most are benign event-level flags: installation logs lose trailing confirmation lines but the agent continues after the shell prompt returns; the heredoc case removes only terminal-echoed text that already appears in the agent's previous message. 
The only behaviorally affected case occurs when the agent inspects a recently written file with \texttt{cat pipeline\_parallel.py}; the compressed view hides part of the file and the agent emits the intended \texttt{[Compression Complaint] need\_full\_output} signal. 
This complaint is used by \agentname{} as feedback to make the corresponding rule more conservative in subsequent use.

\section{Prompt}
\label{prompt}

This appendix presents the prompts used in TACO for task-level rule initialization and rule evolution. As described in the method section, TACO invokes LLMs in three cases: (1) selecting and refining candidate rules during task initialization, (2) generating a new rule for an uncovered high-output command, and (3) revising an overly aggressive rule into a more conservative replacement after an over-compression complaint.

\subsection{Prompt for Task-Level Rule Selection and Refinement}
\label{appendix:prompt_rule_selection}

TACO uses a proposal-stage prompt to initialize the task-specific rule set $R_t$. When historical rules are available in the Global Rule Pool, the model selects, reuses, and refines relevant rules based on the current task context. When no suitable history is available, the model performs cold-start planning and directly proposes an initial rule set from scratch.

\subsubsection*{(a) With-Cache Variant}

This variant is used when the current task category already has historical rules in the Global Rule Pool. The model receives cached rules together with the task description and terminal state, and returns a structured plan containing reused, modified, and newly created rules.

\begin{promptbox}{PROPOSAL PROMPT WITH CACHE}
You are a terminal observation compression strategy expert.

The system already has these baseline filters (you do NOT need to generate rules for these):
- ANSI escape code removal
- System login banner (Ubuntu MOTD) removal
- Empty command polling state handling

Below are historical compression rules from previous tasks. Select the ones
relevant to the current task, modify any that need adjustment, and create
new rules if needed.

Historical rules:
{cached_rules_json}

Current task description:
{instruction}

Task category: {task_category}

Current terminal environment (first 500 chars):
{terminal_state}

Instructions:
1. "selected_rule_ids": List rule_ids of rules to use AS-IS from the historical set
2. "modified_rules": For rules that are close but need adjustment, output the full
   modified rule with a NEW rule_id (e.g., original_id + "_mod")
3. "new_rules": For command types not covered by any historical rule, create new rules

Requirements:
- Only create rules for HIGH-OUTPUT commands (pip, apt, make, pytest, git, docker, etc.)
- Do NOT create rules for short-output commands (ls, cat, echo, pwd, cd)
- NEVER compress error output --- errors must always be fully preserved
- Be conservative: when in doubt, KEEP the line rather than strip it
- Total rules (selected + modified + new) should be 3-7

Output a single JSON object:
{
  "selected_rule_ids": ["id1", "id2"],
  "modified_rules": [
    {
      "rule_id": "string",
      "trigger_regex": "string",
      "description": "string",
      "keep_patterns": ["regex1"],
      "strip_patterns": ["regex1"],
      "keep_first_n": 5,
      "keep_last_n": 10,
      "max_lines": null,
      "summary_header": "[description]",
      "priority": 42
    }
  ],
  "new_rules": [
    {same format as modified_rules}
  ]
}

Output ONLY the JSON object, no other text.
\end{promptbox}

\subsubsection*{(b) Cold-Start Variant}

This variant is used when no suitable historical rules are available for the current task category. Instead of selecting from cached rules, the model predicts which command types are likely to generate long terminal observation and directly constructs an initial rule set.

\begin{promptbox}{PROPOSAL PROMPT NO CACHE}
You are a terminal observation compression strategy expert.

The system already has these baseline filters (you do NOT need to generate rules for these):
- ANSI escape code removal
- System login banner (Ubuntu MOTD) removal
- Empty command polling state handling

Given the task below, predict which terminal commands will produce long outputs,
and create compression rules for them.

Task description:
{instruction}

Task category: {task_category}

Current terminal environment (first 500 chars):
{terminal_state}

Requirements:
- Only create rules for HIGH-OUTPUT commands (pip, apt, make, pytest, git, docker, etc.)
- Do NOT create rules for short-output commands (ls, cat, echo, pwd, cd)
- NEVER compress error output --- errors must always be fully preserved
- Be conservative: when in doubt, KEEP the line rather than strip it
- Generate 3-7 rules

For each rule, provide:
- trigger_regex: regex to match the command string
- description: what this rule does
- keep_patterns: regex patterns for lines that MUST be preserved
- strip_patterns: regex patterns for lines safe to remove
- keep_first_n: always keep first N lines (default 5)
- keep_last_n: always keep last N lines (default 10)
- max_lines: cap on body lines after filtering (null = no cap)
- summary_header: text to show when lines are removed

Output a single JSON object:
{
  "rules": [
    {
      "rule_id": "string",
      "trigger_regex": "string",
      "description": "string",
      "keep_patterns": ["regex1"],
      "strip_patterns": ["regex1"],
      "keep_first_n": 5,
      "keep_last_n": 10,
      "max_lines": null,
      "summary_header": "[description]",
      "priority": 42
    }
  ]
}

Output ONLY the JSON object, no other text.
\end{promptbox}

\subsection{Prompt for New Rule Generation}
\label{appendix:prompt_rule_generation}

This prompt is used when a command produces a long terminal observation but no active rule matches that command type. In this case, TACO treats the output as an uncovered case and asks the model to generate a new compression rule that can be reused in subsequent steps.

\begin{promptbox}{SPAWN NEW PROMPT}
You are a terminal observation compression rule expert.

The agent executed a command that produced a very long observation ({output_length} chars),
but no compression rule exists for this command type.

Command: {command}

Output (first 2000 chars):
{raw_output_head}

Output (last 500 chars):
{raw_output_tail}

Task context: {task_instruction}

Generate a compression rule for this type of command. The rule should:
1. Have a trigger_regex that matches this CATEGORY of command (not just this exact command)
2. Identify repetitive/progress/noise patterns in the output to strip
3. Preserve all error messages, results, and actionable information
4. Be conservative --- when in doubt, keep the line

Output a single JSON object with these fields:
{
  "rule_id": "string",
  "trigger_regex": "string",
  "description": "string",
  "keep_patterns": ["regex1", "regex2"],
  "strip_patterns": ["regex1", "regex2"],
  "keep_first_n": 5,
  "keep_last_n": 10,
  "max_lines": null,
  "summary_header": "[description of what was compressed]",
  "priority": 42
}

Output ONLY the JSON object, no other text.
\end{promptbox}

\subsection{Prompt for Conservative Rule Update After Over-Compression Complaints}
\label{appendix:prompt_rule_update}

This prompt is used when subsequent agent behavior indicates that the compressed output was too aggressive, e.g., when the agent requests the full output, re-executes a command to recover missing details, or otherwise signals that critical information may have been removed. TACO then freezes the triggered rule and asks the model to generate a more conservative replacement.

\begin{promptbox}{SPAWN REPLACEMENT PROMPT}
You are a terminal observation compression rule expert.

The following rule compressed terminal observation too aggressively, causing the
agent to miss critical information.

Old rule (JSON):
{old_rule_json}

Command that was executed: {command}

Original terminal observation (first 2000 chars):
{raw_output_snippet}

Agent's feedback (what it complained about):
{agent_feedback}

Generate a NEW replacement rule that:
1. Keeps the same trigger_regex (targets same command type)
2. Is MORE CONSERVATIVE --- preserves more information
3. Stays SPECIFIC to this command type (don't make a generic "keep everything" rule)
4. Adds the missing information type to keep_patterns
5. Only strips content that is 100% guaranteed noise (progress bars, blank lines, etc.)
6. Uses a new rule_id (suggest: old_id + "_v2" or similar)

Output a single JSON object with these fields:
{
  "rule_id": "string",
  "trigger_regex": "string",
  "description": "string",
  "keep_patterns": ["regex1", "regex2"],
  "strip_patterns": ["regex1", "regex2"],
  "keep_first_n": 5,
  "keep_last_n": 10,
  "max_lines": null,
  "summary_header": "[description of what was compressed]",
  "priority": 42
}

Output ONLY the JSON object, no other text.
\end{promptbox}

\subsection{Prompts for LLM-based Static Baselines}
\label{appendix:prompt_static_baselines}

\subsubsection*{Prompt for LLM-Gen Rules}
\begin{promptbox}{LLM-GEN RULES PROMPT}
You are a terminal observation compression strategy expert.

The system already has these baseline filters (you do NOT need to generate rules for these):
- ANSI escape code removal
- System login banner (Ubuntu MOTD) removal
- Empty command polling state handling

Given the task below, predict which terminal commands will produce long outputs,
and create compression rules for them.

Task description:
{instruction}

Task category: {task_category}

Current terminal environment (first 500 chars):
{terminal_state}

Requirements:
- Only create rules for HIGH-OUTPUT commands (pip, apt, make, pytest, git, docker, etc.)
- Do NOT create rules for short-output commands (ls, cat, echo, pwd, cd)
- NEVER compress error output --- errors must always be fully preserved
- Be conservative: when in doubt, KEEP the line rather than strip it
- Generate 3-7 rules

For each rule, provide:
- trigger_regex: regex to match the command string
- description: what this rule does
- keep_patterns: regex patterns for lines that MUST be preserved
- strip_patterns: regex patterns for lines safe to remove
- keep_first_n: always keep first N lines (default 5)
- keep_last_n: always keep last N lines (default 10)
- max_lines: cap on body lines after filtering (null = no cap)
- summary_header: text to show when lines are removed

Output a single JSON object:
{
  "rules": [
    {
      "rule_id": "string",
      "trigger_regex": "string",
      "description": "string",
      "keep_patterns": ["regex1"],
      "strip_patterns": ["regex1"],
      "keep_first_n": 5,
      "keep_last_n": 10,
      "max_lines": null,
      "summary_header": "[description]",
      "priority": 42
    }
  ]
}

Output ONLY the JSON object, no other text.
\end{promptbox}

\subsubsection*{Prompt for LLM Summarization}

\begin{promptbox}{DEFAULT TERMINAL OUTPUT COMPRESSION PROMPT}
# Role
You are the Terminal Output Safe Compressor, responsible for compressing redundant terminal output without affecting AI Agent decision-making.

# Core Principle (MUST strictly follow)
Content can ONLY be compressed when the following condition is met:
"If this content is compressed from the context, the Agent will still make the exact same correct decision, and the task result will remain unchanged."

# Task Context (IMPORTANT --- Use this to understand what information is critical)
{task_instruction}

# Input Format
1. TASK: The task the Agent is working on (shown above)
2. COMMAND: The shell command executed by the Agent
3. RAW_OUTPUT: The raw terminal output

# Safe Compression Rules

## Content that CAN be safely compressed
1. Progress bars and download statistics (percentage, speed, ETA)
2. Git transfer statistics (object enumeration, compression numbers)
3. System banners and copyright notices (Ubuntu welcome, MOTD)
4. Repetitive log lines with same pattern
5. ANSI color codes and escape sequences

## Content that MUST NEVER be compressed
1. Any error messages --- preserve completely
2. Actual command output results (ls, cat, head, tail output)
3. Interactive prompts (yes/no, password prompts)
4. Path and filename information
5. Version numbers and package names
6. Test results (passed/failed counts)
7. Port numbers, URLs, IP addresses
8. Analysis command output --- CRITICAL: Output from these commands must be preserved COMPLETELY:
   - diff, cmp, comm --- Every line shows critical differences
   - hexdump, xxd, od --- Every byte matters for binary analysis
   - cat -A, cat -v, cat -e --- Shows invisible characters that are crucial
   - strings, objdump, readelf --- Binary inspection results
   - strace, ltrace --- System call traces
   - md5sum, sha*sum --- Checksum values must be exact
9. Program execution results --- When running ./program or similar:
   - ALWAYS preserve: The final output/result (numbers, calculation results, "success"/"failed")
   - ALWAYS preserve: Debug output like "values[x] = y", "result = z"
   - ALWAYS preserve: Any single-line output that could be the program's answer
   - CAN compress: Progress bars (0%...100%), repetitive status updates
   - Example: Debug: x=5, y=10\n15  ->  Keep "Debug: x=5, y=10\n15" (the "15" is the result!)

# Output Format (Strict JSON)
{
  "is_safe_to_compress": boolean,
  "has_error": boolean,
  "status": "success" | "failed" | "running",
  "summary": "string"
}

# Command
{command}

# Terminal Output
{terminal_output}
\end{promptbox}

To evaluate the actual performance stability and verify the effectiveness of the convergence metric (Sec~\ref{Convergence Metric Validation}), we plotted the rolling standard deviation of task accuracy in Fig.~\ref{fig:convergence}(b). The detailed calculation procedure is as follows:

Task accuracy is defined as the percentage of tasks successfully resolved (i.e., verifier reward equals $1$) in a given run. We compute the sample standard deviation of these task accuracies over a sliding window of three consecutive runs ($W=3$). For reference, the baseline variance plotted as horizontal dotted lines in Fig.~\ref{fig:convergence}(b) represents the standard deviation of accuracies across independent baseline runs of the same model without self-evolution.

\newpage

\end{document}